\documentclass[10pt,twocolumn,letterpaper]{article}

\usepackage[pagenumbers]{cvpr} 

\usepackage{graphicx}
\usepackage{amsmath}
\usepackage{amssymb}
\usepackage{booktabs}
\usepackage{dsfont}

\usepackage{times}
\usepackage{epsfig}

\usepackage{setspace}
\usepackage{bm}
\usepackage{multirow}
\usepackage{pifont}
\usepackage{balance}
\usepackage{makecell}
\usepackage{colortbl}
\usepackage{cuted} 
\usepackage{pifont}

\usepackage{multicol}
\usepackage{adjustbox}
\definecolor{mygray}{gray}{.9}
\usepackage{wrapfig}
\usepackage{framed}

\usepackage{color}

\def\paperID{***} 
\def\confName{CVPR}
\def\confYear{2024}

\definecolor{c1}{RGB}{130,60,120}
\definecolor{c2}{RGB}{223,139,111}
\definecolor{c3}{RGB}{144,59,28}
\usepackage{xcolor}
\usepackage[normalem]{ulem} 
\newcommand\hl{\bgroup\markoverwith
  {\textcolor{yellow}{\rule[-.5ex]{2pt}{2.5ex}}}\ULon}

\usepackage[pagebackref,breaklinks,colorlinks]{hyperref}

\usepackage[capitalize]{cleveref}
\crefname{section}{Sec.}{Secs.}
\Crefname{section}{Section}{Sections}
\Crefname{table}{Table}{Tables}
\crefname{table}{Tab.}{Tabs.}

\begin{document}

\title{LLaFS: When Large Language Models Meet Few-Shot Segmentation}
\author{
    Lanyun Zhu\textsuperscript{\rm1} ~~~
    Tianrun Chen\textsuperscript{\rm2} ~~~
    Deyi Ji\textsuperscript{\rm3} ~~~
    Jieping Ye\textsuperscript{\rm3} ~~~ 
    Jun Liu\textsuperscript{\rm1}\thanks{Corresponding Author} ~~~ \\
    Singapore University of Technology and Design \textsuperscript{\rm1} ~~~
    Zhejiang University \textsuperscript{\rm2} ~~~
    Alibaba Group \textsuperscript{\rm3}\\
}

\maketitle

\begin{abstract}
\vspace{-0.5\baselineskip}
This paper proposes LLaFS, the first attempt to leverage large language models (LLMs) in few-shot segmentation. In contrast to the conventional few-shot segmentation methods that only rely on the limited and biased information from the annotated support images, LLaFS leverages the vast prior knowledge gained by LLM as an effective supplement and directly uses the LLM to segment images in a few-shot manner. To enable the text-based LLM to handle image-related tasks, we carefully design an input instruction that allows the LLM to produce segmentation results represented as polygons, and propose a region-attribute table to simulate the human visual mechanism and provide multi-modal guidance. We also synthesize pseudo samples and use curriculum learning for pretraining to augment data and achieve better optimization. LLaFS achieves state-of-the-art results on multiple datasets, showing the potential of using LLMs for few-shot computer vision tasks. 
\end{abstract}
\vspace{-1\baselineskip}

\section{Introduction}
\vspace{-0.5\baselineskip}
Image segmentation is a fundamental task in computer vision with extensive applications. The development of deep learning algorithms \cite{He_2016_CVPR, dosovitskiy2020image, zhu2023learning, foo2023aigenerated, zhu2024ibd} trained on large-scale datasets has brought significant advancements to this field \cite{pspnet, deeplabv3, cheng2022masked, xie2021segformer, Chen_2023_ICCV, Zhu_2023_CVPR}. However, annotating pixel-level segmentation ground truth on a large scale is extremely resource-intensive. Therefore, a more source-efficient learning strategy, few-shot segmentation, has received much attention from academia and holds immense practical value.

In few-shot segmentation, the model should develop category-specific segmentation capabilities based on only a small amount of annotated data, called support images. To achieve this, existing few-shot segmentation methods \cite{okazawa2022interclass, tian2020prior, li2021adaptive, min2021hypercorrelation, zhang2019canet, zhang2019pyramid, nguyen2019feature, liu2021harmonic} typically adopt a support-feature-guided framework. In this framework, relevant features of the target category are extracted from annotated support images and used as guiding information to segment query images. To achieve higher performance, researchers have proposed many methods to explore better ways for support feature extraction \cite{li2021adaptive, okazawa2022interclass, min2021hypercorrelation} and query segmentation assistance \cite{huang2023prototypical, wang2023rethinking, xu2023self}. Although these efforts have demonstrated some success, their segmentation performance is still far from satisfactory. This is because the very limited number of support images contain only a small, incomplete, and biased set of information, so the framework that relies solely on these support-based features for query segmentation inherently suffers from information constraints and cannot achieve a sufficiently high level of accuracy. Therefore, we believe that the further advancement of few-shot segmentation urgently requires an entirely new framework, which should be capable of utilizing richer and more comprehensive information, thereby breaking through the existing framework's bottlenecks to reach better results.

We discover that recent advances in large language models (LLMs) \cite{brown2020language, touvron2023llama, zhao2023survey} can offer potential opportunities to achieve this goal. Specifically, LLMs pre-trained on large-scale corpora have accumulated a vast amount of prior knowledge, which can effectively supplement the insufficient information in support images, thereby resulting in the more effective guidance. Moreover, LLMs have shown to be effective few-shot learners in the field of NLP \cite{brown2020language}. This naturally inspires us to further extend their capabilities to few-shot tasks in other modalities. Based on these insights, we hereby innovatively employ LLMs to tackle few-shot segmentation and introduce an entirely new framework named LLaFS. Unlike some previous segmentation methods that also use language models (LMs) but only for auxiliary purposes, such as utilizing LMs to extract intermediate features \cite{yang2023mianet, he2023primitive, Liu_2023_ICCV_1} or to generate attribute prompts \cite{ma2023open}, our LLaFS directly employs LLMs to produce segmentation results. This makes LMs no longer work as only auxiliary tools, but fully unlock their complete potential in handling the complex computer vision tasks in an end-to-end manner. In this way, we provide an important exploration towards a unified framework that allows LLMs to tackle few-shot tasks in other modalities beyond NLP.

We find that integrating LLM to few-shot segmentation is non-trivial as we face three critical technical challenges: 1) \textit{How to enable the text-based LLM to comprehend and address an image processing task?} 2) \textit{How to leverage both the visual information from support images and the text information from the LLM to guide the query segmentation?} and 3) \textit{How to effectively train the model with only limited data?} To address the first challenge, we draw inspiration from instruction tuning \cite{peng2023instruction} and introduce a task instruction, which is used to explicitly define the few-shot segmentation task within the input of the LLM. To tackle the second challenge, we treat support images as in-context demonstration samples and design a region-attribute corresponding table to extract fine-refined multi-modal guidance information. For the third challenge, we further propose a pseudo-sample synthesis method to augment the pretraining samples and introduce a curriculum learning mechanism to achieve better optimization. By incorporating these designs, our LLaFS can handle few-shot segmentation effectively. We conduct experiments on multiple datasets and achieve state-of-the-art (SOTA) results that significantly outperform existing methods. 

In summary, the contributions of this work are as follows: 1) We introduce LLaFS, the first framework to address few-shot segmentation using large language models. 2) We propose various innovative designs to make better use of LLMs in few-shot segmentation, including a task-tailored instruction, a fine-grained in-context instruction serving as multimodal guidance, and a pseudo-sample-based curriculum pretraining mechanism. 3) Our approach achieves state-of-the-art performance on multiple datasets.

\vspace{-0.5\baselineskip}
\section{Related Work}
\vspace{-0.5\baselineskip}
\noindent \textbf{Few-Shot Segmentation. }To address the issue of conventional semantic segmentation methods \cite{deeplabv3, xie2021segformer, zhu2021learning, cheng2021per, wang2023fvp, sstkd, urur, cagcn, fu2022panoptic, zang2024resmatch} that require a large number of training samples, the task of few-shot segmentation (FSS), which allows to segment a query image using only a small number of annotated support images, has been proposed and gained significant attention \cite{wang2019panet, yang2021mining, liu2020crnet, jiaomask, liu2020part, boudiaf2021few, siam2019amp, lang2022learning, okazawa2022interclass}. Current FSS methods typically adopt a prototype-guided approach \cite{fan2022self, lang2022learning, yang2021mining, nguyen2019feature, li2021adaptive, huang2023prototypical}. They use masked average pooling (MAP) to extract global \cite{fan2022self, nguyen2019feature, liu2022intermediate} or local \cite{li2021adaptive} average prototypes from the backbone features of support images, and then employ these prototypes for guiding the segmentation of query images through feature fusion \cite{li2021adaptive, lang2022learning, liu2022intermediate}, distance computation \cite{min2021hypercorrelation, hong2022cost}, or attention mechanisms \cite{peng2023hierarchical}. However, these methods can only leverage a limited amount of information extracted from a very small number of support images, thus potentially leading to suboptimal results and reduced robustness. To overcome this limitation, \cite{yang2023mianet} uses the more comprehensive word embedding as the general class information to assist in segmentation. Despite some improvements, \cite{yang2023mianet} is still constrained by the limited capabilities of small language models and has not delved deeper into how to better integrate textual information and support image data to achieve more effective guidance. In this paper, we are the first to employ large language models (LLMs) to achieve FSS by using our carefully designed instructions that contain a more effective multimodal guidance. Furthermore, we utilize the LLM to directly produce segmentation results, rather than merely using intermediate features as done in \cite{yang2023mianet}. This offers a brand-new paradigm to FSS.

\noindent \textbf{Large Language Models. }The advent of large language models (LLMs) such as GPT \cite{brown2020language} and Llama \cite{touvron2023llama} has marked the beginning of a new era in artificial intelligence. Thanks to their significantly increased model parameters and training data, these LLMs contain rich prior knowledge and can be efficiently finetuned for specific tasks or application requirements through methods such as prompts \cite{wang2022multitask, liu2022p}, adapters \cite{karimi2021compacter, houlsby2019parameter} and LoRA \cite{hu2021lora}. Recently, researchers have started exploring visual large language models \cite{li2023blip, liu2023llava, wang2023visionllm, su2023pandagpt} to establish a unified framework for multimodal data processing, aiming to override the restriction of LLMs being solely applicable to language data. However, none of them are designed for few-shot tasks in computer vision. In this paper, we introduce the first visual LLM framework for handling few-shot segmentation. To achieve this, we draw inspiration from instruction tuning \cite{peng2023instruction} and in-context learning \cite{min2022rethinking, garg2022can}, and carefully design a suitable form of instruction and demonstration examples tailored for few-shot segmentation. By doing so, our method can enable the LLM to comprehend image data and perform few-shot segmentation effectively.

\begin{figure*}
    \centering
    \includegraphics[width=1\linewidth]{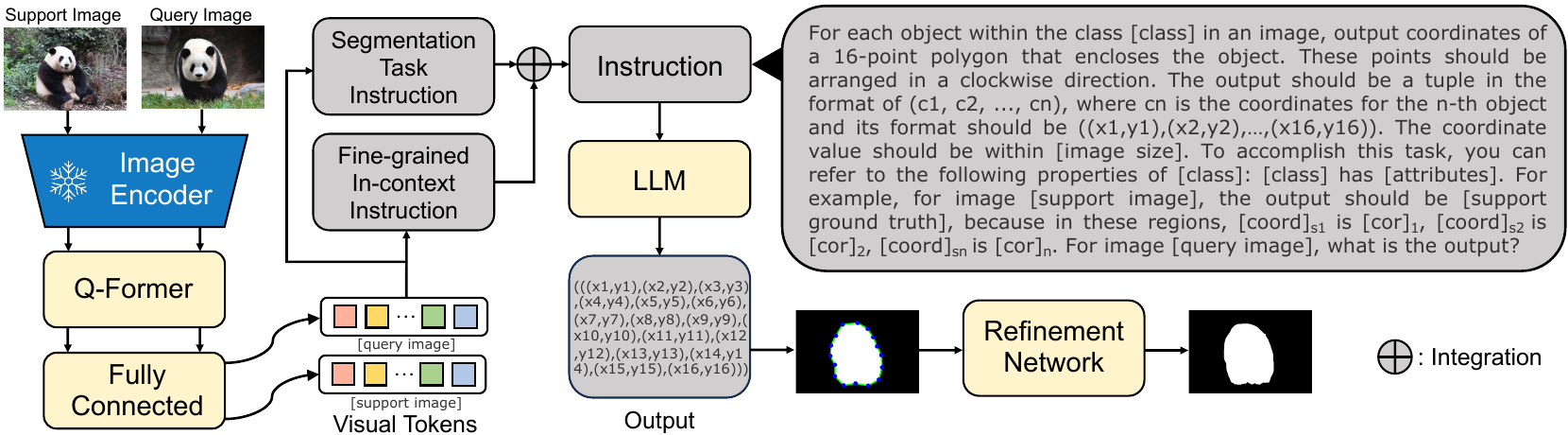}
    \caption{\textbf{Overview of LLaFS.} The image encoder and Q-former extract image features and generate a set of visual tokens. Subsequently, a segmentation task instruction and fine-grained in-context introduction are introduced to provide detailed and comprehensive information. These two instructions are integrated and fed into the LLM to produce the vertices coordinates of polygons that enclose the target object. The segmentation mask represented by this polygon is processed by a refinement network to get the final result.}
    \label{main_fig}
\vspace{-0.8\baselineskip}
\end{figure*}

\vspace{-0.5\baselineskip}
\section{Method}
\vspace{-0.5\baselineskip}
\subsection{Overview}
\vspace{-0.5\baselineskip}
This paper aims to construct an LLM-based framework for few-shot segmentation, i.e, to segment a query image $I_{q}$ based on $K$ support images $\{I_{s}^{k}\}_{k=1}^{K}$ and their ground truth maps $\{G_{s}^{k}\}_{k=1}^{K}$.\footnote{For simplify of illustration, we introduce LLaFS under the one-shot setting. Supp presents how to extend LLaFS to the multi-shot setting.} As shown in Fig.\ref{main_fig}, the overall framework of LLaFS can be divided into three key components: (1) a feature extractor that extracts image features and generates visual tokens; (2) a task-tailored instruction that combines visual tokens, target categories, and task requirements to provide task-related information and support guidance; and (3) an LLM that predicts segmentation masks based on the input instruction, followed by a refinement network to optimize the results. For the feature extractor, we adopt the approach in Blip2 \cite{li2023blip} by using an image encoder followed by a Q-former and a fully-connected layer to generate a set of visual tokens. We use ResNet50 as the image encoder and keep it frozen during training. For the instruction, we carefully design it as the combination of two parts: segmentation task instruction (Sec.\ref{segmentation_task_instruction}) and fine-grained in-context instruction (Sec.\ref{fine_grained_instruction}) to provide comprehensive and detailed guidance. For the LLM, we employ CodeLlama \cite{roziere2023code} with 7 billion parameters that have been fine-tuned through instruction tuning. Note that compared to vanilla Llama, we empirically find that CodeLlama fine-tuned with code generation datasets exhibits higher accuracy and stability in generating structured information like the segmentation result in our task. We equip CodeLlama with LoRA for fine-tuning. All these components work together within the LLaFS framework to achieve high-quality few-shot segmentation. 

\vspace{-0.3\baselineskip}
\subsection{Instruction}
\vspace{-0.3\baselineskip}
As the input of LLM, the instruction is the most crucial component in our framework that makes LLM possible to handle few-shot segmentation. To provide comprehensive information, we design two instructions, namely segmentation task instruction and fine-grained in-context instruction, to respectively provide the LLM with detailed task definitions and fine-grained multi-modal guidance. These two instructions are integrated to formulate the complete instruction as shown in Fig.\ref{main_fig}. In the following Sec.\ref{segmentation_task_instruction} and Sec.\ref{fine_grained_instruction}, we introduce these two instructions in detail. 

\vspace{-0.7\baselineskip}
\subsubsection{Segmentation Task Instruction}\label{segmentation_task_instruction}
\vspace{-0.3\baselineskip}
The LLMs trained on massive text contents have gained strong reasoning capabilities and a vast amount of world knowledge. Language instructions have shown to be a powerful tool for leveraging these knowledge and capability to handle complex tasks \cite{peng2023instruction}. To achieve better results, the instructions need to be sufficiently clear and detailed, whereas those using only simple terminologies such as `performing image segmentation' are evidently too abstract for LLMs to comprehend. Thus, we design a structured instruction to explicitly provide more task details such as the expected input and output formats of few-shot segmentation. Specifically, in our instruction, we represent the pixel-wise segmentation output as a 16-point polygon that encloses the target object \cite{liu2023polyformer}. Note that it is hard for LLMs to directly generate pixel-wise image masks due to LLM's limited number of output tokens. Our alternative solution of generating polygon vertices provides a token-efficient method for using LLMs to achieve pixel-level segmentation.

Furthermore, training solely on text contents makes LLMs difficult to comprehend visual information precisely, especially in our few-shot image segmentation task, where the number of available training images is very scarce. For this, inspired by the success of in-context learning in NLP \cite{min2022rethinking, garg2022can}, we further propose to encode the support image and its ground truth as a visual demonstration example, using it as an intuitive reference in the instruction to teach LLM how to segment a particular class within an image. 
 
By incorporating these designs, we write our segmentation task instruction as: \textcolor{c1}{For each object within the class [class] in an image, output coordinates of a 16-point polygon that encloses the object. These points should be arranged in a clockwise direction. The output should be a tuple in the format of (c1, c2, ..., cn), where cn is the coordinates for the n-th object and its format should be ((x1,y1),(x2,y2),...,(x16,y16)). The coordinate value should be within [image size]. For example, for image [support image], the output should be [support ground truth].} Here, [support image] is the visual tokens from the support image.

\vspace{-0.7\baselineskip}
\subsubsection{Fine-grained In-context Instruction}\label{fine_grained_instruction}
\vspace{-0.5\baselineskip}
\noindent \textbf{Motivation.} The above task instruction makes segmenting a class possible by leveraging LLM's knowledge of the class. In the instruction, the class to be segmented is indicated by the [class] token, which is typically a single noun. However, considering that LLMs have never been trained on images, it is challenging for them to directly align this abstract noun with an image region that may possess a complex internal structure. To address this issue, we drew inspiration from human brains and found that when classifying an unseen new class, the human cognitive system follows a mechanism of `\textit{from general to detailed, from abstract to concrete}' \cite{wisniewski1998relations, murphy1985role}. Specifically, given an unseen class represented by a \textit{general} noun, the human brain first decomposes it into \textit{detailed} attributes based on the acquired knowledge. For example, in the case of an unseen class `owl', a person can first gather information from references to learn about the owl's attributes such as `large round eyes' and `hooked beak'. Subsequently, it can search the image for \textit{concrete} regions that match these \textit{abstract} attributes to determine the presence of the class. 

Inspired by this, we propose a fine-grained in-context instruction to simulate such a human cognitive mechanism based on the support images. For this, we first use ChatGPT to extract detailed attributes of the target class, then we search for regions in support images that correspond to these attributes and generate a corresponding table accordingly. The obtained attributes and table constitute an in-context instruction that is fed into the LLM, which serves as a demonstration example to guide the LLM on how to recognize an image class in a more human-like and fine-grained manner. This alleviates the challenge that LLM cannot perform segmentation well when only inputted with an abstract class noun. Lastly, we introduce an expert-guide framework that refines the instruction to increase its class representation ability. The following sections explain how to generate and refine this instruction in detail.\\

\begin{figure}
    \centering
    \includegraphics[width=0.9\linewidth]{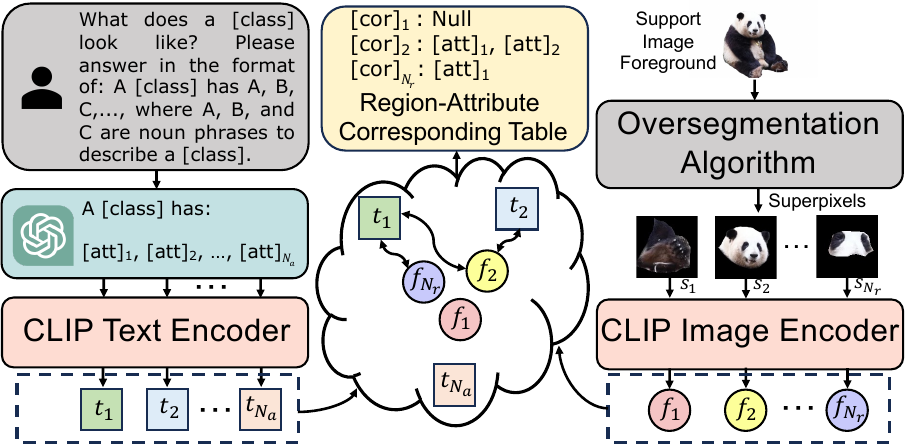}
    \caption{Illustration of how to construct the region-attribute corresponding table used in the fine-grained in-context instruction.}
    \label{fig_2}
\vspace{-0.7\baselineskip}
\end{figure}

\begin{figure}
    \centering
    \includegraphics[width=0.9\linewidth]{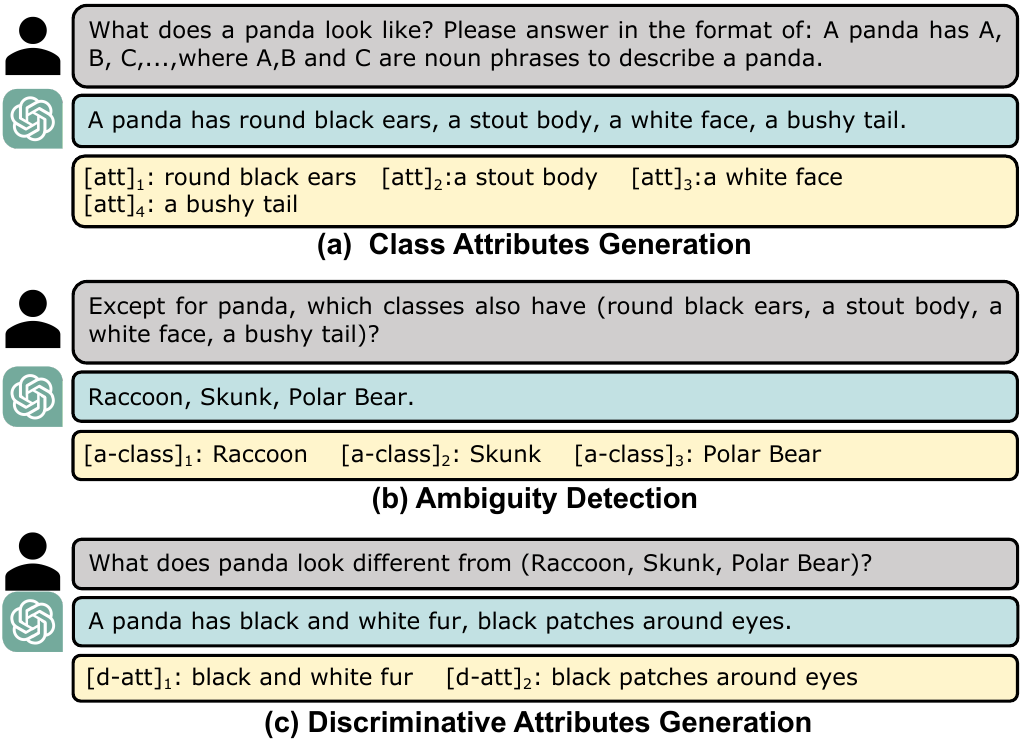}
    \caption{Examples of using ChatGPT for (a) class attributes generation, (b) ambiguity detection and (c) discriminative attributes generation.}
    \label{gpt}
\vspace{-0.5\baselineskip}
\end{figure}

\vspace{-0.5\baselineskip} \noindent \textbf{Attributes Generation} We first simulate the step of `\textit{from general to detailed}' to extract class attributes. Specifically, as shown in Fig.\ref{gpt}(a), we construct a prompt `What does a [class] look like? Please answer in the format of: A [class] has A, B, C,..., where A, B, and C are noun phrases to describe a [class].', and utilize ChatGPT to extract phrases-based attributes that describe the fine-grained details of this class. These attributes are denoted as [attributes] = \{[att]$_{i}$\}$_{i=1}^{N_{a}}$. For each [att]$_{i}$, we utilize `A photo of [att]$_{i}$' as a prompt to extract an embedding $t_{i}$ from the CLIP's text encoder. In this way, we get $\{t_{i}\}_{i=1}^{N_{a}}$ from \{[att]$_{i}$\}$_{i=1}^{N_{a}}$. \\

\vspace{-0.5\baselineskip} \noindent \textbf{Region-attribute Corresponding Table.} After that, we simulate the second step of `\textit{from abstract to concrete}'. To implement this, as shown in Fig.\ref{fig_2}, we propose a region-attribute corresponding table to find the alignment between support image regions and class attributes. For this, we first divide the support foreground into multiple local regions. Specifically, for each object in the support image within the target class, we employ the method in selective search \cite{uijlings2013selective} to generate a set of superpixels $\{s_{i}\}_{i=1}^{N_{r}}$ with different scales in an unsupervised manner. Each $s_{i}$ aggregates pixels that are close in position and similar in features, so it can represent a local region with a specific semantic meaning. Based on each $s_{i}$, a masked image is generated and passed through CLIP's\footnote{We use CLIP fintuned through the method in \cite{liang2023open} to ensure high-quality region-text alignment.} image encoder to produce a feature $f_{i}$. We calculate the cosine similarity between $f_{i}$ and the embedding $t_{j}$ for each attribute [att]$_{j}$, and utilize a thresholding process to establish region-attribute correspondence. Formally,
\begin{equation} \label{get_label}
    {\rm [cor]}_{i} = 
\left[{\rm [att]}_{j}\enspace {\rm for}\enspace j\in [1,N_{a}]\enspace {\rm if}\enspace {\rm cos}(f_{i}, t_{j}) > \alpha\right]
\end{equation}
where ${\rm cos}$ refers to cosine similarity, $\alpha$ is a pre-defined threshold. The obtained [cor]$_{i}$ contains attributes that align with $s_{i}$. In this way, we get \{[cor]$_{i}\}_{i=1}^{N_{r}}$ from $\{s_{i}\}_{i=1}^{N_{r}}$, which serves as an attribute-region corresponding table that can provide the fine-grained multi-modal reference.\\ 

\vspace{-0.5\baselineskip} \noindent \textbf{Instruction Construction. }We integrate the obtained class attributes \{[att]$_{i}$\}$_{i=1}^{N_{a}}$ and corresponding table \{[cor]$_{i}\}_{i=1}^{N_{r}}$, and write the fine-grained in-context instruction as: \textcolor{c1}{The [class] has [attributes]. For example, in [support image], [coord]$_{s_{1}}$ is [cor]$_{1}$, [coord]$_{s_{2}}$ is [cor]$_{2}$..., [coord]$_{s_{N_{r}}}$ is [cor]$_{N_{r}}$.} Here, [coord]$_{s_{i}}$ is the coordinates of $s_{i}$ in the format of a 16-point polygon. By using this instruction, we provide the LLM with a reference about the attributes of the target class and their corresponding regions in the support image. In this way, a demonstration example can be created that simulates how the human cognitive mechanism recognizes the support image as a class. Using this example as a reference, the LLM can be taught how to understand and segment an image class in a fine-grained manner.\\

\vspace{-0.5\baselineskip} \noindent \textbf{Expert-guide Framework for Instruction Refinement.}
The above-mentioned instruction constructed by the obtained attributes \{[att]$_{i}$\}$_{i=1}^{N_{a}}$ and table \{[cor]$_{i}\}_{i=1}^{N_{r}}$ can be directly input into LLM for guidance. However, due to variations in camera angles and instances of occlusion, not every attribute can be directly matched to a region within the support image. Thus, the obtained table \{[cor]$_{i}\}_{i=1}^{N_{r}}$ may contain only a subset of attributes within \{[att]$_{i}$\}$_{i=1}^{N_{a}}$. Unfortunately, we find the combinations of these partial attributes may be insufficient for the unambiguous recognition of the target class. For example, the combination `wheels, windows, doors' can refer to `train', `car', and `bus' interchangeably. Due to the ambiguous table, the instruction may be misleading. To alleviate this issue, we propose an expert-guide framework to refine the instruction. In this framework, we first employ ChatGPT to identify ambiguous classes of the existing table, then we extract additional attributes that can distinguish the target class from these ambiguous classes for refinement. As long as these additional attributes can be aligned with local regions in the support image, the refined table based on them will become unambiguous. In this way, the class representation ability and comprehensiveness of the instruction can be improved.

Specifically, this framework generates a refined instruction through the following three steps: \textbf{\textit{1) Ambiguity Detection.}} As shown in Fig.\ref{gpt}(b), We employ ChatGPT to identify potential ambiguous classes in the obtained table \{[cor]$_{i}\}_{i=1}^{N_{r}}$. Specifically, we denote the attributes contained in \{[cor]$_{i}\}_{i=1}^{N_{r}}$ as [partial-attributes] and ask ChatGPT `Except for [class], which classes also have [partial-attributes]?'\footnote{Due to space limitations, we omit the description of format control prompts for inputting into ChatGPT. See Supp for details.} In this way, we obtain a set of ambiguous classes denoted as [a-classes]=\{[a-class]$_{i}\}_{i=1}^{N_{ac}}$ from ChatGPT's feedback. \textbf{\textit{2) Discriminative Attributes Generation.}} As shown in Fig.\ref{gpt}(c), To avoid being misled by these ambiguous classes, we use `What does [class] look different from [a-classes]?' as a text prompt, enabling ChatGPT to generate attributes that are more discriminative from these classes. The obtained attributes \{[d-att]$_{i}\}_{i=1}^{N_{d}}$ are added to [attributes] for updating. \textbf{\textit{3) Table and Instruction Refinement.}} Finally, we use these updated attributes to reperform Eq.\ref{get_label} to obtain a refined table. The updated attributes and table are reassembled to form a refined fine-grained in-context instruction.

We found that a single execution of the three steps already resolves ambiguities in over 85\% of the instructions. For the remaining 15\% of instructions, we observe that because the newly-acquired discriminative attributes still couldn't find matching regions in the support image, the resulting table after refinement remains to be ambiguous. Therefore, we iteratively apply the last two steps of this framework until the ambiguity is eradicated. To achieve this, from the second iteration onwards, we replace the text prompt in the discriminative attributes generation step with `Apart from [all-discriminative-attributes], tell me more differences in appearance between [class] and [a-classes]', where [all-discriminative-attributes] refers to the discriminative attributes {[d-att]$_i$} obtained from all previous iterations. By doing so, we enable the iterative framework to continuously discover more discriminative attributes and verify whether they have matched regions in the support image. Eventually, when the framework successfully discovers discriminative attributes [d-att]$_{i}$ that can be aligned with the support image or reaches the maximum number of iterations, we terminate the iteration. For efficiency, we set the maximum number of iterations to 3, in which 96\% of the ambiguities have been entirely eradicated.

\vspace{-0.5\baselineskip}
\subsection{Segmentation Prediction}
\vspace{-0.5\baselineskip}
We integrate segmentation task instruction and fine-grained in-context instruction to formulate the complete instruction as shown in Fig.\ref{main_fig}. With this instruction as input, the LLM can predict the coordinates of a 16-point polygon that surrounds the target object. Finally, to rectify the imprecision caused by the polygon representation of object edges, a refinement network comprising a pixel decoder and a mask transformer is introduced to generate a refined segmentation mask by using the polygon as the initial mask. Please see Supp for the detailed structures of this network. 

\vspace{-0.3\baselineskip}
\subsection{Curriculum Pretraining with Pseudo Samples}
\noindent \textbf{Motivation. }After carefully designing the model structure and instruction format, the next challenge is how to train LLaFS effectively to achieve high segmentation performance. Previous work \cite{liu2023llava} has shown that the success of instruction tuning often relies on extensive training data. However, due to the difficulty of obtaining pixel-annotated labels, the segmentation datasets used for training typically contain only an insufficient number of images. To address this issue, we propose to generate pseudo support-query pairs and use them to pretrain the LLM. The LLM's ability to handle few-shot segmentation can thus be enhanced by seeing more visual samples. \\

\vspace{-0.5\baselineskip} \noindent \textbf{Pseudo Sample Generation. }Specifically, we propose a method to generate pseudo support-query pairs with the following three steps: \textbf{\textit{1) Pseudo foreground-background partition.}} We first use bezier curves to randomly generate a contour within an image region. The area surrounded by this contour is considered as the foreground within the target class, while the regions outside the contour are treated as the background. \textbf{\textit{2) Noise filling for pseudo support generation.}} We fill the foreground with Gaussian noise that has a random mean value. For background, we first randomly divide it into multiple subregions to simulate the complex backgrounds in real images, then we fill each one with a random Gaussian noise that has a mean value distinct from the foreground's noise. The obtained image is used as the support image. \textbf{\textit{3) Pseudo query generation.}} We use a similar approach to generate a query image. Note that in this process, the contour and the mean value of foreground noise are no longer completely randomly determined but adjusted based on those used for generating the support image. This is done to ensure that the foreground regions of both support and query images have similar contour shapes and internal features, so they can reflect the same category. Please refer to Supp for the adjustment details.\\

\begin{figure}
    \centering
    \includegraphics[width=0.85\linewidth]{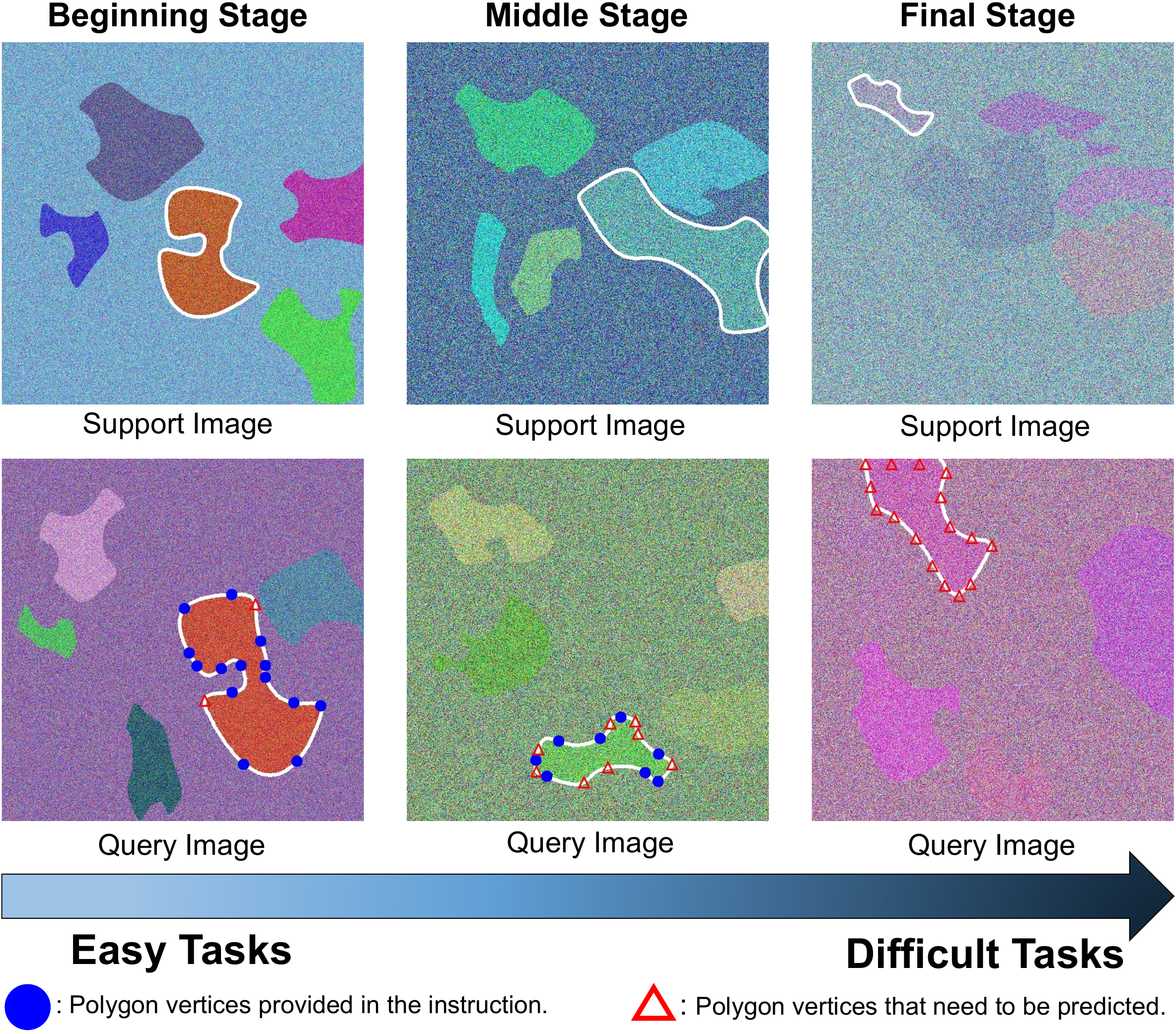}
    \caption{Examples of pseudo samples generated at different pretraining stages. Foreground regions are marked by white contours. As pretraining progresses, pseudo images have reduced intra-image foreground-background differences and greater support-query foreground differences. Meanwhile, the number of polygon vertex coordinates provided in the instruction decreases, while the predicted vertex count increases. These changes gradually increase the pretraining difficulty. (Best viewed in color)}
    \label{fig_3}
\vspace{-0.8\baselineskip}
\end{figure}

\vspace{-0.5\baselineskip} \noindent \textbf{Curriculum Pretraining.} The synthetic support-query pairs can be directly used for pretraining. However, it is observed that this approach will result in a slow convergence rate. A reason for this problem is that the LLM has not been previously trained on image data, so optimizing it to directly grasp a complex image processing task is challenging. To address this issue, we propose a progressive pretraining approach inspired by the success of curriculum learning \cite{wang2021survey}, in which we initiate the model's pretraining with a simple task and gradually increase the task's difficulty until it ultimately reaches the requirements of segmentation. 

Specifically, during pretraining, we incrementally raise the task's difficulty from the following two aspects: \textbf{\textit{1) Image understanding. }}During pretraining, by controlling the difference between mean values of different filled noise, we gradually increase the difference in foreground between support and query, while reducing the internal difference between foreground and background within each image. This makes it more challenging for LLM to perform few-shot guidance and partition foreground-background areas as pretraining progresses. \textbf{\textit{2) Polygon generation. }}We observe that generating a polygon represented by a combination of vertex coordinates is another challenge for the LLM. Therefore, we adopt a progressive strategy here as well. Instead of training the model to directly predict the coordinates of all 16 points of the polygon, we randomly provide the coordinates of $M$ points in the instruction and let the LLM to predict the coordinates of the remaining $16-M$ points. During pretraining, we gradually decrease the value of $M$ from 15 to 0. This means that the model receives fewer hints and is required to predict more vertex coordinates as pretraining progresses. Consequently, the pretraining difficulty gradually increases, ultimately reaching the task of predicting all 16 points for segmentation. Experimental results show that this curriculum learning approach allows the model to converge better and achieve higher results. Please see Supp for more technical details on how we increase the difficulty in image understanding and polygon generation.

Ultimately, the model is trained on the realistic few-shot segmentation dataset after completing the aforementioned pretraining process.

\begin{table*}[t]
\begin{minipage}[t]{0.65\linewidth}
 \centering
 \vspace{0pt}
 \setlength\tabcolsep{2pt}
	\resizebox{1\linewidth}{!}{
		\renewcommand{\arraystretch}{1.03}
		\begin{tabular}{lll|ccccc|ccccc}
			\hline
			\multirow{2}{*}{Dataset} & \multirow{2}{*}{Method} & \multirow{2}{*}{Conference} & \multicolumn{5}{c|}{1-shot}           & \multicolumn{5}{c}{5-shot}            \\ \cline{4-13} 
			&    &   & Fold-0 & Fold-1 & Fold-2 & Fold-3 & Mean  & Fold-0 & Fold-1 & Fold-2 & Fold-3 & Mean  \\ \hline
			\multirow{11}{*}{PASCAL-$5^{i}$} & NTRENet \cite{liu2022learning} & CVPR'22 & 65.4 & 72.3 & 59.4 & 59.8 & 63.2 & 66.2 & 72.8 & 61.7 & 62.2 & 65.7\\
            & BAM\cite{lang2022learning} & CVPR'22 & 69.0 & 73.6 & 67.5 & 61.1 & 67.8 & 70.6 & 75.1 & 70.8 & 67.2 & 70.9 \\
            & AAFormer\cite{wang2022adaptive} & ECCV'22 & 69.1 & 73.3 & 59.1 & 59.2 & 65.2 & 72.5 & 74.7 & 62.0 & 61.3 & 67.6 \\
            & SSP\cite{fan2022self} & ECCV'22 & 60.5 & 67.8 & 66.4 & 51.0 & 61.4 & 67.5 & 72.3 & 75.2 & 62.1 & 69.3\\
            & IPMT\cite{liu2022intermediate} & NeurIPS'22 & 72.8 & 73.7 & 59.2 & 61.6 & 66.8 & 73.1 & 74.7 & 61.6 & 63.4 & 68.2\\
            & ABCNet\cite{wang2023rethinking} & CVPR'23 & 68.8 & 73.4 & 62.3 & 59.5 & 66.0 & 71.7 & 74.2 & 65.4 & 67.0 & 69.6\\
            & HDMNet \cite{peng2023hierarchical}& CVPR'23 & 71.0 & 75.4 & 68.9 & 62.1 & 69.4 & 71.3 & 76.2 & 71.3 & 68.5 & 71.8\\
            & MIANet\cite{yang2023mianet} & CVPR'23 & 68.5 & 75.8 & 67.5 & 63.2 & 68.7 & 70.2 & 77.4 & 70.0 & 68.8 & 71.7\\
            & MSI\cite{moon2023msi} & ICCV'23 & 71.0 & 72.5 & 63.8 & 65.9 & 68.5 & 73.0 & 74.2 & 70.5 & 66.6 & 71.1\\
            & SCCAN\cite{xu2023self} & ICCV'23 & 68.3 & 72.5 & 66.8 & 59.8 & 66.8 & 72.3 & 74.1 & 69.1 & 65.6 & 70.3\\
             \cline{2-13}  
             \rowcolor{mygray}& LLaFS  & - & \color{red}{74.2} & \color{red}{78.8} & \color{red}{72.3} & \color{red}{68.5} & \color{red}{73.5} & \color{red}{75.9} & \color{red}{80.1} & \color{red}{75.8} & \color{red}{70.7} & \color{red}{75.6}\\
            \hline
            \multirow{12}{*}{COCO-$20^{i}$} & NTRENet\cite{liu2022learning} & CVPR'22 & 36.8 & 42.6 & 39.9 & 37.9 & 39.3 & 38.2 & 44.1 & 40.4 & 38.4 & 40.3\\
            & BAM\cite{lang2022learning} & CVPR'22 & 43.4 & 50.6 & 47.5 & 43.4 & 46.2 & 49.3 & 54.2 & 51.6 & 49.6 & 51.2\\
            & SSP\cite{fan2022self} & ECCV'22 & 35.5 & 39.6 & 37.9 & 36.7 & 47.4 & 40.6 & 47.0 & 45.1 & 43.9 & 44.1 \\
            & AAFormer\cite{wang2022adaptive} & ECCV'22 & 39.8 & 44.6 & 40.6 & 41.4 & 41.6 & 42.9 & 50.1 & 45.5 & 49.2 & 46.9\\ 
            & MM-Former\cite{zhang2022mask} & NeurIPS'22 & 40.5 & 47.7 & 45.2 & 43.3 & 44.2 & 44.0 & 52.4 & 47.4 & 50.0 & 48.4\\
            & IPMT\cite{liu2022intermediate} & NeurIPS'22 & 41.4 & 45.1 & 45.6 & 40.0 & 43.0 & 43.5 & 49.7 & 48.7 & 47.9 & 47.5 \\
            & ABCNet\cite{wang2023rethinking} & CVPR'23 & 42.3 & 46.2 & 46.0 & 42.0 & 44.1 & 45.5 & 51.7 & 52.6 & 46.4 & 49.1\\
            & HDMNet \cite{peng2023hierarchical} & CVPR'23 & 43.8 & 55.3 & 51.6 & 49.4 & 50.0 & 50.6 & 61.6 & 55.7 & 56.0 & 56.0\\
            & MIANet\cite{yang2023mianet} & CVPR'23 & 42.5 & 53.0 & 47.8 & 47.4 & 47.7 & 45.8 & 58.2 & 51.3 & 51.9 & 51.7\\
            & MSI\cite{moon2023msi} & ICCV'23 & 42.4 & 49.2 & 49.4 & 46.1 & 46.8 & 47.1 & 54.9 & 54.1 & 51.9 & 52.0\\
            & SCCAN\cite{xu2023self} & ICCV'23 & 40.4 & 49.7 & 49.6 & 45.6 & 46.3 & 47.2 & 57.2 & {59.2} & 52.1 & 53.9\\
             \cline{2-13}  
             \rowcolor{mygray} & LLaFS  & - & \color{red}{47.5} & \color{red}{58.8} & \color{red}{56.2} & \color{red}{53.0} & \color{red}{53.9} & \color{red}{53.2} & \color{red}{63.8} & \color{red}{63.1} & \color{red}{60.0} & \color{red}{60.0}\\
            \hline
	\end{tabular}}
 \vspace{-0.7\baselineskip}
  \caption{Performance comparison with other methods on PASCAL-5$^i$ and COCO-$20^{i}$.}
  \label{comp_sota}
  \end{minipage}
  \begin{minipage}[t]{0.35\linewidth}
    \vspace{0pt}
    \centering
    \resizebox{1\linewidth}{!}{
    \centering
    \setlength\tabcolsep{25pt}
    \renewcommand{\arraystretch}{1}
    \begin{tabular}{l | c }
    \toprule
    Method & mIoU\\
    \midrule
    LLaFS & 74.2 \\
    \midrule
    LLaFS w/o support images & 56.9\\
    \bottomrule
    \end{tabular}
    }
    \vspace{-0.7\baselineskip}
      \caption{Effectiveness of support images.}
    \label{effect_ablation}
    \vspace{0.4\baselineskip}
    \resizebox{1\linewidth}{!}{
    \centering
    \setlength\tabcolsep{8pt}
    \renewcommand{\arraystretch}{1}
    \begin{tabular}{l | c }
    \toprule
    Method & mIoU\\
    \midrule
    LLaFS & 74.2 \\
    \midrule
    LLaFS w/o class attributes & 70.7\\
    LLaFS w/o region-attribute corresponding table & 69.7\\
    LLaFS w/o instruction refinement & 70.6\\
    LLaFS w/o iterative refinement &71.8\\
    \bottomrule
    \end{tabular}
    }
     \vspace{-0.7\baselineskip}
    \caption{Ablation results of fine-grained in-context instruction. }
    \label{ablation_fine}
     \vspace{0.3\baselineskip}
    \setlength\tabcolsep{2pt}

    \begin{minipage}[b]{1\linewidth}
           \includegraphics[width=0.95\linewidth]{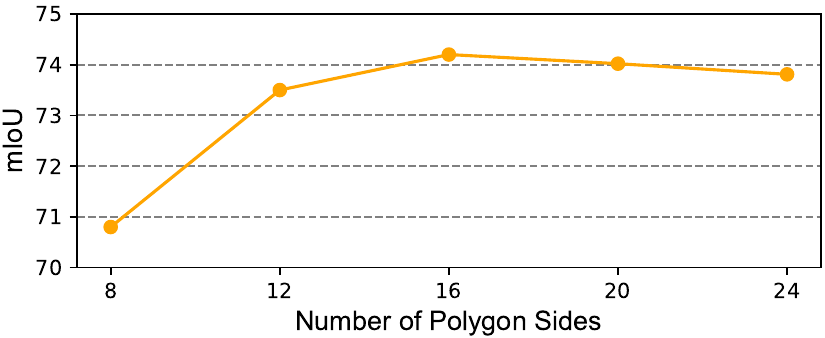} 
    \vspace{-1\baselineskip}
     \makeatletter\def\@captype{figure}\makeatother\caption{Ablation for the number of polygons' sides.}
    \label{ablation_number}
    \end{minipage}

    \end{minipage}
 \vspace{-1\baselineskip}
  \end{table*}

\vspace{-0.9\baselineskip}
\section{Experiments}
\vspace{-0.5\baselineskip}
\subsection{Implementation Details and Training Settings}\label{implementation}
\vspace{-0.5\baselineskip}
We set the number of queries in the Q-former to 100 and the threshold $\alpha$ in Eq.\ref{get_label} to 0.2. The ground truth of polygon vertices is obtained in polar coordinates \cite{xie2020polarmask}. Specifically, starting from the object center, 16 rays are uniformly emitted at equal angular intervals $\triangle \theta=22.5^{\circ}$. The points of intersection between these rays and the object contour are taken as the ground truth of the polygon vertices. More implementation details about pseudo sample generation and curriculum pretraining are presented in Supp.

The overall model is trained in three stages. In the first stage, we freeze the LLM, pretrain the Q-former and fully-connected layers for 100K steps using the datasets\footnote{COCO is excluded from the pretraining set to avoid test data leakage.} and methods in Blip2 \cite{li2023blip}. In the second stage, we freeze the Q-former, equip the LLM with LoRA, and pretrain the fully-connected layers, LLM and refinement network using the pseudo-sample-based curriculum learning method for 60k steps. In the third stage, we train the fully-connected layers, LLM and refinement network on the realistic few-shot segmentation dataset (25 epochs for PASCAL-5$^{i}$ and 3 epochs for COCO-20$^{i}$). AdamW \cite{loshchilov2018decoupled} is used as the optimizer with the cosine annealing schedule and an initial learning rate of 0.0002. The model is trained on 16 A100 GPUs.

\begin{table}[t]
    \centering
    \begin{adjustbox}{width=1.0\columnwidth,center}
    \renewcommand{\arraystretch}{1.0}
    \setlength\tabcolsep{7pt}
    \begin{tabular}{l | c}
    \toprule
    Method & mIoU \\
    \midrule
    LLaFS & 74.2\\
    \midrule
    LLaFs w/o segmentation task instruction w/ abstract summary & 67.7\\
    LLaFS w/o fine-grained in-context instruction & 67.0\\
    LLaFS w/o refinement network & 69.1\\
    LLaFS w/o pseudo-sample-based curriculum pretraining & 63.5\\
     \bottomrule
    \end{tabular}
    \end{adjustbox}
    \vspace{-0.5\baselineskip}
    \caption{Effectiveness of different components in our LLaFS.}
    \label{ablation_component}
\vspace{-0.7\baselineskip}
\end{table}

\vspace{-0.5\baselineskip}
\subsection{Comparison with State-of-the-arts}
\vspace{-0.5\baselineskip}
Table.\ref{comp_sota} presents the comparisons with other few-shot segmentation methods on two datasets: PASCAL-5$^{i}$ and COCO-20$^{i}$. Following previous papers, we use different folds as the test set, with the remaining folds utilized for training. This approach yields four sets of experimental results along with their mean result. Across all datasets and experimental settings, our method consistently outperforms others and achieves a remarkably significant improvement compared to previous state-of-the-art results. For instance, on PASCAL-5$^{i}$, LLaFS improves mIoUs by 4.1\% and 3.8\% in the 1-shot and 5-shot settings respectively compared to the second-ranking method. Notably, our approach still exhibits great advantages on the more complex and challenging COCO-20$^{i}$ dataset, with increases of 3.9\% and 4.0\% in the 1-shot and 5-shot settings respectively. This could be attributed to the rich prior knowledge embedded in the LLM and our carefully designed instructions, which enable the models to handle complex images effectively and robustly. These results demonstrate the outstanding performance of our method and highlight the huge potentiality of using LLMs for tackling few-shot segmentation tasks.

\vspace{-0.5\baselineskip}
\subsection{Ablation Study}
\vspace{-0.5\baselineskip}
Using the 1-shot setting of the PASCAL-5$^{i}$ dataset, we perform several ablation studies to evaluate different components and designs in our method. More ablation studies are presented in \textbf{Supp}.

\noindent \textbf{Effectiveness of Different Components}
We validate the effectiveness of key components in our method, including (1) the segmentation task instruction, (2) fine-grained in-context instruction, (3) the refinement network, and (4) the pseudo-sample-based curriculum pretraining. Experimental results are presented in Table.\ref{ablation_component}. Replacing the detailed segmentation task introduction with an abstract summary `perform image segmentation' decreases mIoU by 6.5\%. Not using the other components individually decreases the mIoU by 7.2\%, 5.1\%, and 10.7\%, respectively. These results demonstrate the importance and effectiveness of each component in our approach.

\noindent \textbf{Effectiveness of Support Images. }In the instruction, the annotated support images provide the LLM with crucial visual guidance. As shown in Table.\ref{effect_ablation}, if we do not use the support image as a demonstration example, mIoU decreases by 17.3\%. This highlights the importance of the support image as a few-shot sample in our approach, demonstrating that our LLsFS benefits not solely from LLM's prior knowledge in an open-vocabulary manner but indeed gains further improvement from the provided few-shot samples.

\noindent \textbf{Number of Polygon's Sides. }In the segmentation task instruction, we represent the segmentation output mask as a region enclosed by a 16-point polygon. We find that the number $M$ of sides in the polygon is also a factor affecting the model's performance. Therefore, we test the relationship between mIoU and $M$ and present the results in Fig.\ref{ablation_number}. We observe that when $M$ is small, the model's performance is suboptimal. This is because polygons with a smaller number of sides cannot accurately describe object edges. As $M$ increases, the mIoU gradually improves. However, when $M$ exceeds 16, we observe a slight decrease in performance. This could be because a larger $M$ increases the task's complexity for LLM to tackle. Based on the results, we chose $M=16$ as our setting.

\noindent \textbf{Ablation of Fine-grained In-context Instruction.} Table.\ref{ablation_fine} presents the evaluation of different components and designs in our fine-grained in-context instruction, including (1) class attributes, (2) attribute-region corresponding table, (3) expert-guide framework for instruction refinement, and (4) iterative refinement. We observe that when these components are removed, mIoU decreases by 3.5\%, 4.5\%, 3.6\% and 2.4\%, respectively. These results demonstrate the rationality of our designs in this instruction and their effectiveness in improving performance.

\begin{table}[t]
    \centering
    \begin{adjustbox}{width=0.8\columnwidth,center}
    \renewcommand{\arraystretch}{1.0}
    \setlength\tabcolsep{20pt}
    \begin{tabular}{l | c}
    \toprule
    Method & mIoU \\
    \midrule
    LLaFS & 74.2\\
    \midrule
    LLaFS w/o pseudo samples & 63.5\\
    LLaFS w/o curriculum strategy & 67.3\\
    LLaFS w/ random pseudo query generation & 63.9\\
     \bottomrule
    \end{tabular}
    \end{adjustbox}
    \vspace{-0.8\baselineskip}
    \caption{Ablation of pseudo-sample-based curriculum pretraining.}
    \label{ablation_pesudo}
\vspace{-0.3\baselineskip}
\end{table}

\begin{figure}
    \centering
    \includegraphics[width=0.8\linewidth]{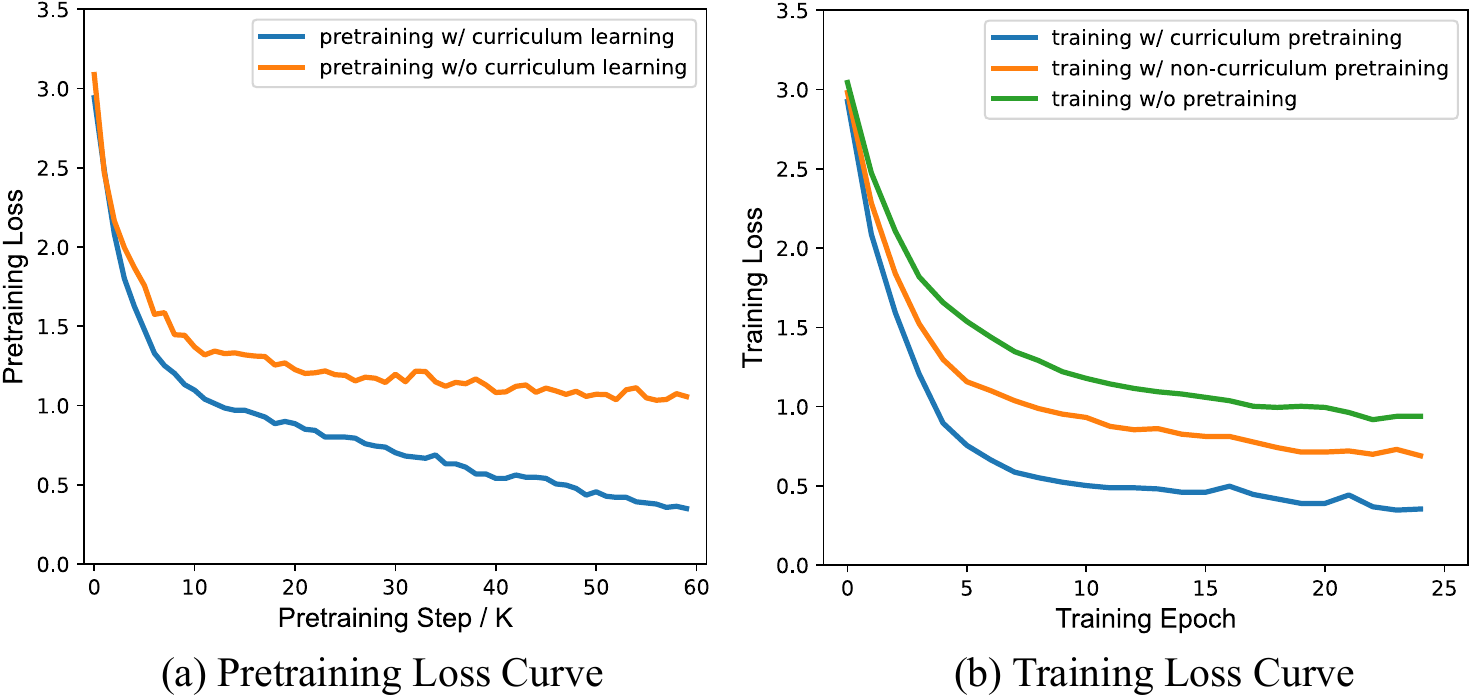}
    \vspace{-1\baselineskip}
    \caption{Pretraining (a) and training (b) loss curves in different settings. Curriculum pretraining results in the best convergence in both pretraining and training stages. (Best viewed in color)}
    \vspace{-0.7\baselineskip}
    \label{loss}
\end{figure}

\noindent \textbf{Ablation of Pseudo-sample-based Curriculum Pretraining.} We further evaluate the key techniques in our pseudo-sample-based curriculum pretraining mechanism and the results are presented in Table.\ref{ablation_pesudo}. (1) When we do not employ pseudo-samples for pretraining, mIoU decreases by 10.7\%. (2) Removing the curriculum training strategy that gradually increases the training task difficulty reduces mIoU by 6.9\%. (3) When generating pseudo support-query samples, to ensure that the support and query can reflect the same category, the contour and the mean value of foreground noise used to generate the query image are adjusted based on those used for generating the support image. When this strategy is not employed and random generation is used instead, mIoU decreases by 10.3\%. These results demonstrate the effectiveness of our method's designs. 

\vspace{-0.5\baselineskip}
\subsection{Loss Curves}
\vspace{-0.5\baselineskip}
In Fig.\ref{loss}, we present the loss curves during the pretraining and training stages. We observe from Fig.\ref{loss}(a) that without the use of curriculum learning, the pretraining task becomes excessively challenging, which causes the model optimization to quickly reach a bottleneck with difficulties in convergence. After using our curriculum learning mechanism, the model achieves significantly better convergence. Furthermore, in Fig.\ref{loss}(b), we compare the loss reduction during the training stage when using curriculum pretraining, non-curriculum pretraining, and not using pretraining. The model without pretraining converges the slowest, while the model with curriculum pretraining converges the fastest during the training stage. These results demonstrate the effectiveness of our curriculum-learning-based pretraining approach in enhancing the model's convergence speed.

\begin{figure}
    \centering
    \includegraphics[width=0.8\linewidth]{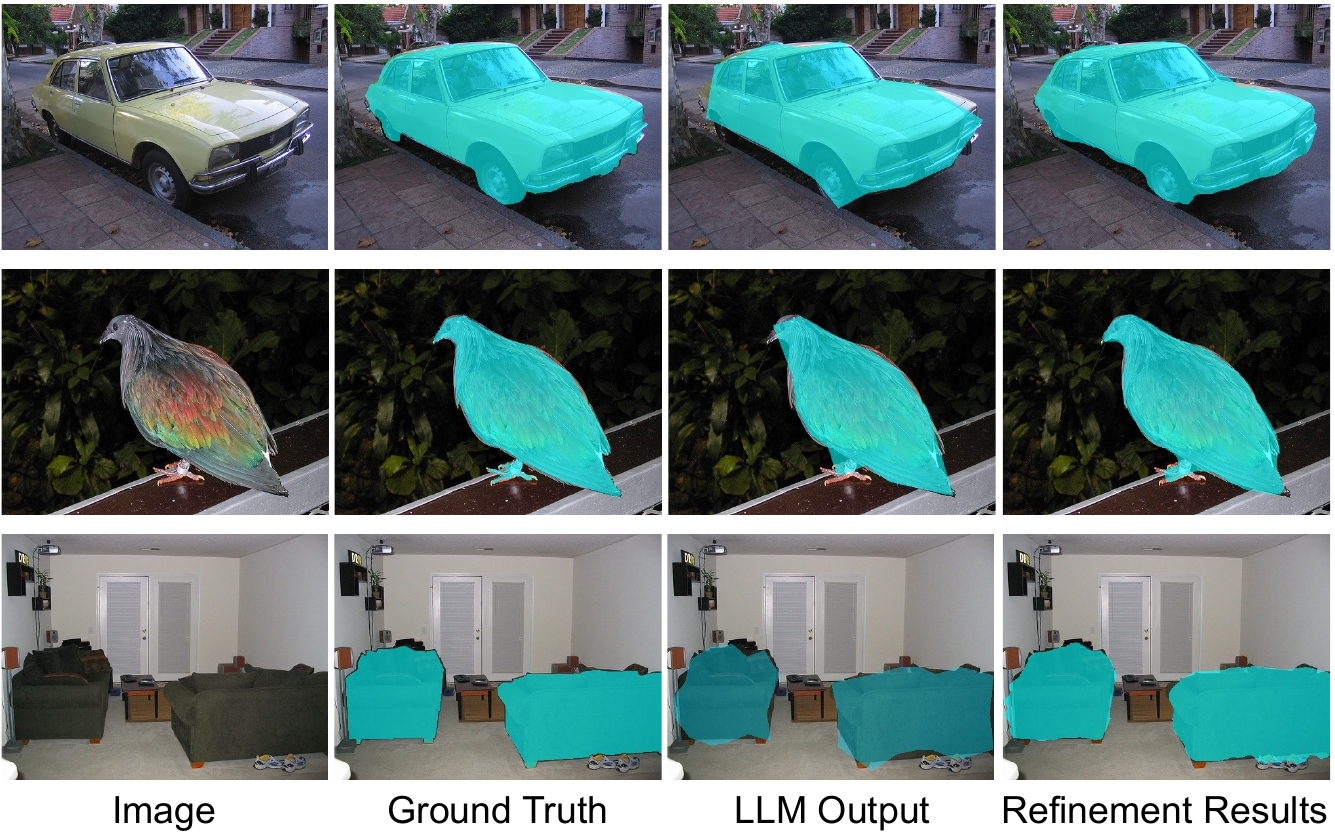}
    \vspace{-1\baselineskip}
    \caption{Visualization of segmentation results.}
    \label{visual}
\vspace{-0.7\baselineskip}
\end{figure}

\vspace{-0.5\baselineskip}
\subsection{Visualization}
\vspace{-0.5\baselineskip}
In Fig.\ref{visual}, we present visual samples of the segmentation results from LLaFS, including the image, ground truth, the LLM output, and the results after refinement. It can be observed that the polygons output by LLM already achieve good segmentation performance, and the results after refinement are more accurate, particularly in the object's edge regions. It is worth noting that, as shown in the third row of Fig.\ref{visual}, our method still achieves high-performance segmentation when there are more than one target object in the image. These results demonstrate the excellent performance of LLaFS.

\vspace{-0.5\baselineskip}
\section{Conclusion}
\vspace{-0.5\baselineskip}
This paper proposes LLaFS, a novel framework that for the first time, leverages large language models (LLMs) to address few-shot segmentation in an end-to-end manner. To enable LLMs to perform this visual task, we introduce a segmentation task instruction to provide detailed task definitions, and propose a fine-grained in-context instruction to simulate human cognitive mechanisms and provide refined multimodal reference information. We also propose a pseudo-sample-based curriculum pretraining mechanism to augment the training samples required for instruction tuning. Our extensive experiments demonstrate the effectiveness of LLaFS, which achieves significantly superior state-of-the-art results across multiple datasets and settings. We consider LLaFS as an important exploration towards an LLM framework capable of addressing few-shot tasks in different modalities beyond natural language processing.  
\noindent \textbf{Acknowledgement}
This research is supported by the National Research Foundation, Singapore under its AI Singapore Programme (AISG Award No: AISG2-PhD-2021-08-006), MOE AcRF Tier 2 projects MOE-T2EP20222-0009 and MOE-T2EP20123-0014.

{\small
\bibliographystyle{ieee_fullname}
\bibliography{egbib}
}

\clearpage

\appendix
The supplementary materials are arranged as follows: In Sec.\ref{method_details}, we illustrate additional method details of the LLaFS framework. In Sec.\ref{more_imlemntation}, we provide more implementation details of our method. In Sec.\ref{more_experiments}, we present more experimental results to further demonstrate the effectiveness of our method and designs. In Sec.\ref{example_step}, we showcase some examples of the key steps from input to output of the LLaFS framework. 
\vspace{-0.3\baselineskip}
\section{More Method Details of LLaFS}\label{method_details}
\vspace{-0.3\baselineskip}
\subsection{Complete Input for ChatGPT}
\vspace{-0.3\baselineskip}
In our proposed expert-guide framework for instruction refinement, we input ChatGPT with a text prompt describing the task requirements to achieve ambiguity detection and discriminative attributes generation (see Fig.3 of main paper for examples). In practice, the complete ChatGPT input is formed by combining this text prompt with a \textit{\textbf{format control prompt}} that explicitly specifies the format of the output we expect from ChatGPT. When the output format is unified and fixed, it is easier for us to extract the ambiguous classes \{[a-class]$_{i}\}_{i=1}^{N_{ac}}$ and discriminative attributes \{[d-att]$_{i}\}_{i=1}^{N_{d}}$ from ChatGPT's feedback automatically and efficiently. Specifically, the complete ChatGPT inputs for ambiguity detection and discriminative attributes generation are written as:\\

\vspace{-0.5\baselineskip}
\noindent \textbf{Ambiguity Detection:} Except for [class], which classes also have [partial-attributes]? Please answer in the format of: the following classes also have them: A, B, C, ..., , where A, B and C are the name of classes. If there is no such a class, reply `no'.\\

\vspace{-0.5\baselineskip}
\noindent \textbf{Discriminative Attributes Generation:} What does [class] look different from [a-classes]? Please answer in the format of: [class] has A, B, C,..., where A,B and C are noun phrases to describe the difference of [class] compared to [a-classes].\\

\vspace{-0.5 \baselineskip}
\noindent \textbf{Discriminative Attributes Generation (from the second iteration onwards):} Apart from [all-discriminative-attributes], tell me more differences in appearance between [class] and [a-classes]. Please answer in the format of: [class] has A, B, C,..., where A,B and C are noun phrases to describe more differences of [class] compared to [a-classes] apart from the given ones.

\vspace{-0.3\baselineskip}
\subsection{More Details of Pseudo-sample-based Curriculum Pretraining}
\vspace{-0.3\baselineskip}
\subsubsection{Contour Generation Method}
\vspace{-0.3\baselineskip}
When generating pseudo images, we first randomly generate a contour within an image region, and the area surrounded by this contour is considered as the foreground within the target class. This contour is generated as a Bézier curve constructed by 10 randomly-generated control points. We will release code to illustrate it in detail. 

\subsubsection{Details of Pseudo Support-query Generation}
\vspace{-0.3\baselineskip}
The detailed method for generating a support-query pair can be summarized as the following steps:\\

\vspace{-0.5\baselineskip}
\noindent \textbf{Step 1: Support foreground-background partition.} We first randomly generate a contour within an image region. The area surrounded by this contour is considered as the foreground within the target class, while the regions outside the contour are treated as the background. For the background, we use random contours to divide it into multiple subregions to simulate the diverse backgrounds encountered in real images. The number of subregions is randomly selected from 1 to 5.\\

\vspace{-0.5\baselineskip}
\noindent \textbf{Step 2: Support noise filling.} We randomly generate an array $m_{sf}\in \mathbb{R}^{3}$ within the value range [0, 255], and utilize it as the RGB mean to generate a Gaussian noise for filling the support foreground region. Subsequently, for each subregion of the support background, we randomly generate another array $m_{sb}\in \mathbb{R}^{3}$ as the mean to generate a Gaussian noise for filling this subregion. We constrain the random generation space of $m_{sb}$ to satisfy the distance constraint $||m_{sb} - m_{sf}|| \in [a, b]$, where $a, b$ are two adjustable parameters. By adjusting the values of $a$ and $b$, we can manage the difference between the foreground and background within each synthetic image. Sec.\ref{details_curriculum} illustrates how to adjust them in different pretraining steps. \\

\vspace{-0.5\baselineskip}
\noindent \textbf{Step3: Query foreground-background partition by adjusting from support. }To ensure that the support foreground and query foreground have similar shapes so that they can reflect the same category, the contour used to generate the query foreground is adjusted based on that used for generating the support foreground. Specifically, we first add a standard Gaussian noise to the ten control points that are used to generate the support foreground contour. Subsequently, a noised contour is generated from these noised control points, followed by the random rotation and scaling between [0.5, 1.5] for further adjustment. After that, we randomly place the resulted contour in another position of the image, and the region enclosed by which is regarded as the query foreground. Finally, we use the same approach as the support background to partition the query background region.\\

\vspace{-0.5\baselineskip}
\noindent \textbf{Step4: Query noise filling by adjusting from support. }Using the same method as for the support generation, we randomly generate arrays $m_{qf}\in \mathbb{R}^{3}$ and $m_{qb}\in \mathbb{R}^{3}$ and use them as RGB means to generate Gaussian noises, which are then applied to fill the query foreground and each subregion of the query background. To ensure that the support foreground and query foreground have similar internal features so that they can reflect the same category, we constrain the random generation space of $m_{qf}$ to satisfy the distance constraint $||m_{qf} - m_{sf}|| \in [c, d]$, where $c$ and $d$ are adjustable parameters. For the background's $m_{qb}$, we impose two constraints to determine its random generation space: (1) similar to the support background, we constrain the difference between the query background and query foreground by $||m_{qb} - m_{qf}|| \in [a, b]$. (2) To ensure that query foreground is the most similar region to the support foreground in the query image, we further constrain the difference between the query background and the support foreground to be greater than the difference between the query foreground and the support foreground. This is achieved by constraining $||m_{qb} - m_{sf}|| > ||m_{qf} - m_{sf}||$. Under these constraints, we randomly generate $m_{qf}$ and $m_{qb}$ to serve as the means of noises, which are used to fill different regions to obtain the pseudo query image.

\vspace{-0.5\baselineskip}
\subsubsection{Details of Curriculum Pretraining}\label{details_curriculum}
\vspace{-0.3\baselineskip}
During pretraining, we incrementally raise the task’s difficulty from the following two aspects:\\

\vspace{-0.5\baselineskip}
\noindent \textbf{(1) Image understanding. }During pretraining, by controlling the difference between mean values of different filled noise, we gradually increase the difference in foreground between support and query, while reducing the internal difference between foreground and background within each image. This makes it more challenging for LLM to perform few-shot guidance and partition foreground-background areas as pretraining progresses. We implement this strategy by adjusting the parameters $a, b, c, d$ introduced in the previous section. 

Specifically, the interval $[a,b]$ constrains the difference between the foreground and background within an image. Therefore, during the pretraining process, to reduce this difference, we gradually decrease the values of $a, b$ until $a$ eventually reaches 0. Denoting the total number of pretraining steps as $N_{p}$ ($N_{p}=60K$ in our experiments), the values of $a_{n}$ and $b_{n}$ at step $n$ are formulated as:
\begin{equation}
    \begin{aligned}
        &a_{n} = a_{0} - \frac{n.a_{0}}{N_{p}},\\
        &b_{n} = a_{n} + b_{0} - a_{0},\\
    \end{aligned}
\end{equation}
where $a_{0}$ and $b_{0}$ are the hyper-parameters that define the initial values of $a$ and $b$ in the first step of pretraining. 

The interval $[c, d]$ constrains the difference between the support foreground and query foreground. Therefore, during the pretraining process, to enlarge this difference, we gradually increase the values of $c$ and $d$, making $c$ to be increased from 0 to $c_{N_{p}}$ as the step progresses from 0 to $N_{p}$. In this way, the values of $c_{n}$ and $d_{n}$ at step $n$ are formulated as:
\begin{equation}
    \begin{aligned}
        &c_{n} = \frac{n.c_{N_{p}}}{N_{p}},\\
        &d_{n} = c_{n} + d_{N_{p}} - c_{N_{p}},\\
    \end{aligned}
\end{equation}
where $c_{N_{p}}$ and $d_{N_{p}}$ are the hyper-parameters that define the final values of $a$ and $b$ in the last step of pretraining. 

In this approach, $a_{0}$, $b_{0}$, $c_{N_{p}}$ and $d_{N_{p}}$ are predefined hyper-parameters, which are respectively set to 100, 150, 50, and 100 in our framework. Note that, our experiments demonstrate that \textit{\textbf{the performance of LLaFS is NOT sensitive to these hyper-parameters}}. See Sec.\ref{more_experiments} and Table.\ref{hyper_setting} for details.\\

\vspace{-0.5\baselineskip}
\noindent \textbf{(2) Polygon generation.}
During the pretraining stage, we randomly provide the coordinates of $M$ points in the instruction and let the LLM to predict the coordinates of the remaining $16-M$ points. $M$ is decreased by 1 every $N_{p}/30$ steps in the first half pretraining process with $N_{p}/2$ steps. By doing so, we gradually decrease the value of $M$ from 15 to 0. This means that the model receives fewer hints and is required to predict more vertex coordinates as pretraining progresses. Consequently, the pretraining difficulty gradually increases, ultimately reaching the task of predicting all 16 points for segmentation. In the last half pretraining process, we keep $M=0$ for pretraining.

\vspace{-0.5\baselineskip}
\subsubsection{Instruction in Pretraining.}
\vspace{-0.3\baselineskip}
In the pretraining stage, the instruction for inputting into LLM is written as: \textcolor{c1}{For the target object in a query image that has the same class as the support image foreground, output coordinates of a 16-point polygon that encloses the object. These points should be arranged in a clockwise direction and the format of their coordinates is ((x1,y1),(x2,y2),...,(x16,y16)). The coordinate value should be within [image size]. For support image [pseudo support image], the foreground is [support foreground]. For the target object in the query image [pseudo query image], the output should be [masked-gt]. What is the remaining points?} Here, [masked-gt] refers to retaining part of the ground truth vertices for the 16-point polygon as hints, while replacing the remaining parts to be predicted with a [mask] token as in \cite{fried2022incoder}.

\begin{figure}
    \centering
    \includegraphics[width=0.85\linewidth]{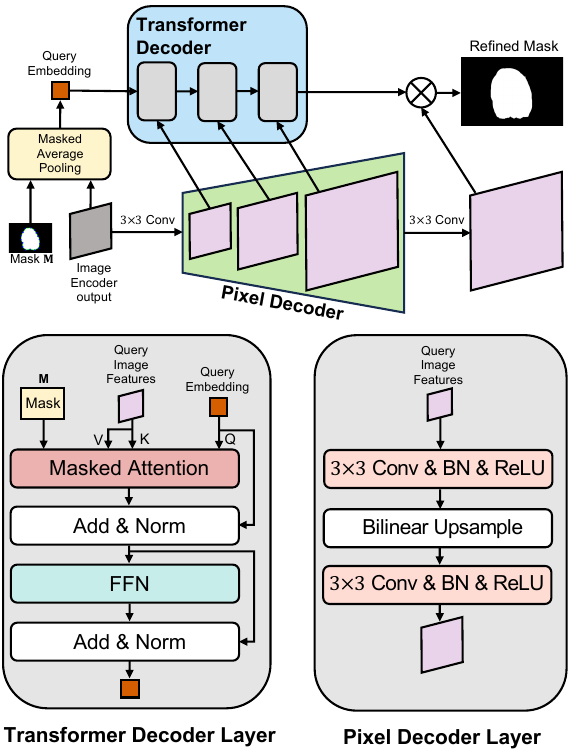}
     \caption{Structure of the refinement network. This network is lightweight, comprising only 6 convolution layers and 3 attention layers. }
    \vspace{-0.7\baselineskip}
    \label{fig_refine_net}
\vspace{-0.3\baselineskip}
\end{figure}

\subsection{Refinement Network}
With the instruction as input, the LLM can predict the coordinates of a 16-point polygon. We use 1 to fill the area enclosed by the polygon and 0 to fill the area outside the polygon. In this way, we obtain a binary segmentation mask denoted as $\textbf{M}$. To rectify the imprecision caused by the polygon representation of object edges, we further introduce a refinement network to obtain a more refined segmentation result. As shown in Fig.\ref{fig_refine_net}, this refinement network follows a similar structure to Mask2Former \cite{cheng2022masked}, comprising a pixel decoder that progressively increases the sizes of query image feature maps and a masked transformer decoder for optimizing the queries. $\textbf{M}$ is used as the mask for the masked attention in the transformer decoder. Readers can refer to Sec3.21 of \cite{cheng2022masked} for more details of masked attention. 

Note that, compared to the vanilla Mask2Former, our method does not employ a heavy transformer to construct the pixel decoder; instead, we use a simple structure composed of a small number of convolution and bilinear upsampling layers. Moreover, we do not iteratively apply the transformer decoder but use it only once. These modifications reduce the computational complexity, making our refinement network to be very lightweight with only 6 convolution layers and 3 attention layers. As the output from the LLM already achieves pretty good results, using this lightweight network for refinement is completely sufficient.

\vspace{-0.3\baselineskip}
\subsection{Expanding LLM Vocabulary with Coordinate Tokens}
\vspace{-0.3\baselineskip}
We expand the vocabulary of LLM by adding 384 coordinate tokens denoted as ([c-0], [c-1], ..., [c-383]), where [c-i] represents the coordinate value $i$. This design makes our model more efficient, requiring fewer tokens for the input and output of LLM.

\vspace{-0.3\baselineskip}
\subsection{Loss Function}
\vspace{-0.3\baselineskip}
Using the Vertex Coordinates outputted by the LLM to construct a binary mask as the input for the refinement network is a non-differentiable process. Therefore, we employ two separate loss functions to train the refinement network and the remaining components of LLaFS, respectively. In this way, the overall loss can be written as:
\begin{equation}
    \mathcal{L} = \mathcal{L}_{llm} + \mathcal{L}_{ref},
\end{equation}
where $\mathcal{L}_{llm}$ denotes the loss for training fully-connected layers and LLM, $\mathcal{L}_{ref}$ denotes the loss for training the refinement network. For $\mathcal{L}_{llm}$, we first use bipartite matching to align the LLM-predicted outputs with each object in the ground truth, then we use cross-entropy loss to compute $\mathcal{L}_{llm}$. For $\mathcal{L}_{ref}$, we use cross-entropy loss with online hard examples mining (OHEM) strategy to compute it. 

\subsection{Extension to Multi-shot Setting}
In the main paper, we introduce LLaFS under the one-shot setting. The method for the multi-shot setting can be easily extended from the one-shot method. Specifically, for each support image, we extract a set of visual tokens and a region-attribute corresponding table using the method illustrated in main paper. These pieces of information from all $K$ support images are then incorporated into the following instruction for feeding into the LLM: \textcolor{c1}{For each object within the class [class] in an image, output coordinates of a 16-point polygon that encloses the object. These points should be arranged in a clockwise direction. The output should be a tuple in the format of (c1, c2, ..., cn), where cn is the coordinates for the n-th object and its format should be ((x1,y1),(x2,y2),…,(x16,y16)). The coordinate value should be within [image size]. To accomplish this task, you can refer to the following properties of [class]: [class] has [attributes]. For example, for image [support image 1], the output should be [support ground truth 1], because in these regions, [coord]$_{s_{1}}$ is [cor]$_{1}$, [coord]$_{s_{2}}$ is [cor]$_{2}$..., [coord]$_{s_{N_{r}}}$ is [cor]$_{N_{r}}$; ...; for image [support image K], the output should be [support ground truth K], because in these regions, [coord]$_{s_{1}}$ is [cor]$_{1}$, [coord]$_{s_{2}}$ is [cor]$_{2}$..., [coord]$_{s_{N_{r}}}$ is [cor]$_{N_{r}}$. For image [query image], what is the output?}

\begin{table*}[t]
\centering
\vspace{0pt}
{
    \centering
    \begin{minipage}{0.25\linewidth}
    \begin{adjustbox}{width=1.0\columnwidth,center}
    \renewcommand{\arraystretch}{1.0}
    \setlength\tabcolsep{3pt}
    \begin{tabular}{l | c c}
    \toprule
    Method & PASCAL-5$^{i}$ & COCO-20$^{i}$\\
    \midrule
    Painter \cite{wang2023images} & 64.5 & 32.8\\
    SegGPT \cite{wang2023seggpt} & 83.2 & 56.1\\
    \midrule
    LLaFS & 88.3 & 64.2\\
     \bottomrule
    \end{tabular}
    \end{adjustbox}
    \subcaption{Comparison with SegGPT and Painter.}
    \label{comp_seggpt}
    \end{minipage}
}
\hspace{0.05in}
{
    \centering
    \begin{minipage}{0.15\linewidth}
    \begin{adjustbox}{width=1.0\columnwidth,center}
    \renewcommand{\arraystretch}{1.5}
    \setlength\tabcolsep{3pt}
    \begin{tabular}{l | c}
    \toprule
    Method & mIoU\\
    \midrule
    LLaFS & 74.2\\
    \midrule
    LLaFS w/o LLM & 62.3\\
     \bottomrule
    \end{tabular}
    \end{adjustbox}
    \subcaption{Effectiveness of large language models.}
    \label{effect_llm}
    \end{minipage}
}
\hspace{0.05in}
{
    \centering
    \begin{minipage}{0.15\linewidth}
    \vspace{9pt}
    \begin{adjustbox}{width=1.0\columnwidth,center}
    \renewcommand{\arraystretch}{1.5}
    \setlength\tabcolsep{6pt}
    \begin{tabular}{l | c}
    \toprule
    LLM & mIoU\\
    \midrule
    Llama2 & 69.8\\
    Code Llama & 74.2\\
     \bottomrule
    \end{tabular}
    \end{adjustbox}
    \subcaption{Llama VS Code Llama as the LLM of LLaFS.}
    \label{comp_llm}
    \end{minipage}
}
\hspace{0.05in}
{
    \centering
    \begin{minipage}{0.37\linewidth}
    \vspace{-9.5pt}
    \begin{adjustbox}{width=1.0\columnwidth,center}
    \renewcommand{\arraystretch}{1.5}
    \setlength\tabcolsep{7pt}
    \begin{tabular}{l | c}
    \toprule
    Method & mIoU\\
    \midrule
    LLaFS & 74.2\\
    \midrule
    LLaFS w/ curriculum polygon generation in training & 74.6\\
     \bottomrule
    \end{tabular}
    \end{adjustbox}
    \subcaption{Curriculum polygon generation in the training stage.}
    \label{abaltion_cu_train}
    \end{minipage}
}
\hspace{0.1in}
{
    \begin{minipage}{0.4\linewidth}
    \vspace{9pt}
    \centering
    \begin{adjustbox}{width=1\columnwidth, center}
    \renewcommand{\arraystretch}{1.0}
    \setlength\tabcolsep{10pt}
    \begin{tabular}{l | c}
    \toprule
    Method & mIoU\\
    \midrule
    Our Curriculum Strategy & 74.2\\
    \midrule
    Masking with a Fixed Ratio $\lambda=0.25$ & 69.5\\
    Masking with a Fixed Ratio $\lambda=0.5$ & 71.5\\
    Masking with a Fixed Ratio $\lambda=0.75$ & 71.1\\
     \bottomrule
    \end{tabular}
    \end{adjustbox}
    \subcaption{Curriculum strategy vs fixed-ratio masking for polygon generation.}
    \label{abaltion_fr}
    \end{minipage}
}
\hspace{0.2in}
{
    \centering
    \begin{minipage}{0.55\linewidth}
    \vspace{9pt}
    \begin{adjustbox}{width=1.0\columnwidth,center}
    \renewcommand{\arraystretch}{0.88}
    \setlength\tabcolsep{20pt}
    \begin{tabular}{l | c}
    \toprule
    Method & mIoU\\
    \midrule
    LLaFS & 74.2\\
    \midrule
    LLaFS w/o curriculum strategy in image understanding & 72.0\\
    LLaFS w/o curriculum strategy in polygon generation & 68.1\\
    \midrule
    LLaFS w/o increasing SF-QF difference & 72.9\\
    LLaFS w/o reducing F-B difference & 72.6\\
     \bottomrule
    \end{tabular}
    \end{adjustbox}
    \subcaption{Ablation results of curriculum pretraining. SF, QF, F, B refer to support foreground, query foreground, foreground, background. }
    \label{abaltion_cu}
    \end{minipage}
}
\vspace{-0.3\baselineskip}
\caption{\textbf{More experimental results.}
}
\vspace{-0.7\baselineskip}
\label{more_exp}
\end{table*}

\vspace{-0.3\baselineskip}
\section{More Implementation Details}\label{more_imlemntation}
\vspace{-0.3\baselineskip}
\noindent \textbf{Image Encoder. }ResNet50 from the CLIP is used as the image encoder. In ResNet50, the output features from stage 3 and stage 4 are resized to 1/8 of the input size and concatenated with the output features from stage 2. This combined feature is used as the input for the Q-former and the pixel decoder in the refinement network.\\

\vspace{-0.5\baselineskip}
\noindent \textbf{Q-former. }The Q-former has 8 layers with the dimension of 384. The input for the text transformer in the Q-former is `a photo of [class].'.\\

\vspace{-0.5\baselineskip}
\noindent \textbf{Refinement Network. }Feature dimension in the refinement network is 128.\\

\vspace{-0.5\baselineskip}
\noindent \textbf{Generation of Region-attribute Corresponding Table. }Generating region-attribute corresponding table requires additional time due to the use of ChatGPT. To prevent this additional computation from affecting training efficiency, we pre-generate the table for each image before training and include it as part of the dataset that can be directly used in all experiments. We will release these tables to facilitate future research.\\

\vspace{-0.5\baselineskip}
\noindent \textbf{Data Augmentation.} We employ random horizontal flipping, random noise padding, random cropping, and random resizing for data augmentation. Note that to prevent the random cropping from causing the mismatch between the region-attribute corresponding table and the augmented support image, we constrain the range of random cropping when augmenting each support image to ensure that the foreground region within the target class is not cropped.\\

\vspace{-0.5\baselineskip}
\noindent \textbf{Other Training Settings. }The batch size is 32. The input image size is (384, 384).

\vspace{-0.5\baselineskip}
\section{More Experimental Results}\label{more_experiments}
\vspace{-0.3\baselineskip}

\noindent \textbf{Comparison with SegGPT and Painter.}
In addition to LLaFS, SegGPT \cite{wang2023seggpt} and Painter \cite{wang2023images} can also achieve few-shot segmentation through in-context learning. Fig.\ref{comp_seggpt} presents the comparison results with these methods. For a fair comparison, we follow SegGPT by combining different segmentation datasets for training and allow the categories in training to cover the categories in testing. It is observed that on both PASCAL-5$^{i}$ and COCO-20$^{i}$, our approach can achieve significant advantages. These results demonstrate that our LLaFS can perform in-context-based few-shot segmentation more effectively, which is benefited from the rich prior knowledge contained in LLM and our carefully-designed fine-grained multi-modal demonstration examples.\\

\vspace{-0.5\baselineskip}
\noindent \textbf{Effectiveness of Large Language Models}
Thanks to the rich prior knowledge and powerful few-shot capabilities, large language models (LLM) play a crucial role in ensuring the high effectiveness of our LLaFS framework. To demonstrate this, we remove the LLM from LLaFS and validate the effectiveness of a few-shot segmentation model that is composed of the remaining parts of LLaFS. 

Specifically, to ensure that the few-shot segmentation can be performed by only using the remaining components, we make the following modifications: (1) When extracting visual tokens from the support image using the Q-former, the vanilla cross-attention that interacts the learned queries with support image features is replaced with the masked attention as in \cite{cheng2022masked}. This change ensures that the obtained support tokens are only related to the foreground region where the target category is located. (2) After the Q-former, we add a cross-attention to interact support tokens with query tokens. This allows query tokens to perceive reference information from the support foreground. The query tokens obtained through this step are used as the input query embeddings for the transformer decoder in the refinement network, which produces the final segmentation result.

As shown in Table.\ref{effect_llm}, although the other network components remain largely unchanged, removing LLM significantly decreases mIoU by 11.9\%. This demonstrates the crucial role of the LLM in our LLaFS framework.\\

\vspace{-0.5\baselineskip}
\noindent \textbf{Llama VS Code Llama.} In LLaFS, we employ Code Llama instead of the vanilla Llama as the large language model. As shown in Table.\ref{comp_llm}, the performance of using Code Llama is 4.4\% better than using Llama. This improvement could be attributed to the fact that Code Llama has been fine-tuned on the code generation dataset, so it is more skilled in generating structured data with fixed formats, such as the segmentation results in our task.\\

\vspace{-0.5\baselineskip}
\noindent \textbf{Ablation of Curriculum Pretraining. } In the curriculum pretraining strategy, we gradually increase the difficulty of the pretraining tasks from two aspects: (1) image understanding and (2) polygon generation. As shown in Table.\ref{abaltion_cu}, Not applying the curriculum strategy in each of these two aspects decreases the mIoU by 2.2\% and 6.1\%, respectively. To increase the difficulty of image understanding, we employ two methods when synthesizing support-query pairs: (1) increasing the difference between support foreground and query foreground, and (2) reducing the difference between foreground and background within each image. Removing each of these two methods decreases the mIoU by 1.3\% and 1.6\%, respectively. These results demonstrate the effectiveness of our designs in the approach.\\

\vspace{-0.5\baselineskip}
\noindent \textbf{Curriculum Strategy VS Fixed-Ratio Masking for Polygon Generation. }We further test a masked strategy used in \cite{he2022masked, wang2023images, wang2023seggpt}, in which we randomly provide a fixed ratio $\lambda$ of vertex coordinates in the instruction, and let the model predict the remaining vertices. 
We test three values for $\lambda$: 0.25, 0.5, and 0.75. As shown in Table.\ref{abaltion_fr}, the performances of all these methods are worse than our curriculum strategy. These results demonstrate the effectiveness of our approach, showing the importance of dynamically increasing the learning difficulty during the pretraining process.\\

\vspace{-0.5\baselineskip}
\noindent \textbf{Curriculum Polygon Generation in the Training Stage. }In addition to employing the curriculum polygon generation on synthetic images during the pretraining stage, we also test the further usage of this strategy to realistic data during the training stage. As shown in Table.\ref{abaltion_cu_train}, we observe that such a modification cannot significantly improve performance. One possible reason could be that the model has acquired sufficient ability to generate 16-point coordinates through pretraining on the pseudo samples, so it no longer requires the continued use of this strategy in the subsequent training stage.\\

\begin{table}[t]
    \centering
    \begin{adjustbox}{width=1.0\columnwidth,center}
    \renewcommand{\arraystretch}{1.0}
    \setlength\tabcolsep{50pt}
    \begin{tabular}{l | c}
    \toprule
    $(a_{0}, b_{0}, c_{N_{p}}, d_{N_{p}})$ & mIoU\\
    \midrule
    (100, 150, 50, 100) & 74.2\\
    (75, 125, 75, 125) & 74.0\\
    (125, 175, 25, 75) & 74.3\\
    (100, 150, 75, 125) & 74.0\\
    (75, 125, 50, 100) & 74.0\\
     \bottomrule
    \end{tabular}
    \end{adjustbox}
    \vspace{-0.5\baselineskip}
    \caption{Different settings for hyper-parameters $(a_{0}, b_{0}, c_{N_{p}}, d_{N_{p}})$ of the pseudo-sample-based curriculum pretraining mechanism.}
    \label{hyper_setting}
\vspace{-0.7\baselineskip}
\end{table}

\vspace{-0.5\baselineskip}
\noindent \textbf{Hyper-parameter Settings for Pseudo-sample-based Curriculum Pretraining. }As discussed in detail in Sec.\ref{details_curriculum}, our proposed pseudo-sample-based curriculum pretraining involves four hyper-parameters $(a_{0}, b_{0}, c_{N_{p}}, d_{N_{p}})$. The results for different combinations of these hyper-parameters are presented in Table.\ref{hyper_setting}. It can be observed that our method can consistently achieve excellent and similar results across different $(a_{0}, b_{0}, c_{N_{p}}, d_{N_{p}})$ settings. These results demonstrate that the performance of LLaFS is NOT sensitive to these hyper-parameters.\\

\begin{figure}
    \centering
    \includegraphics[width=0.92\linewidth]{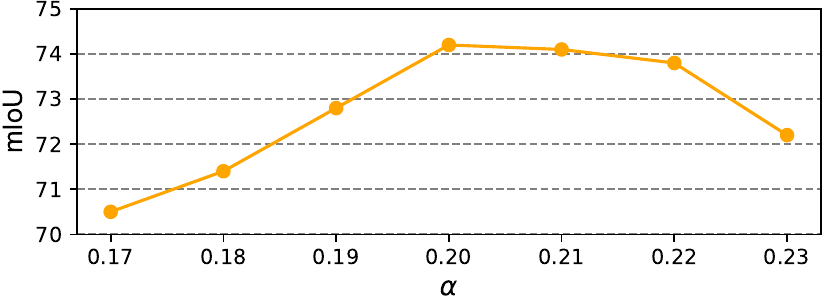}
     \caption{Using different values of threshold $\alpha$ in Eq.1 of main paper. }
    \vspace{-0.7\baselineskip}
    \label{alpha}
\vspace{-0.3\baselineskip}
\end{figure}

\vspace{-0.5\baselineskip}
\noindent \textbf{Ablation for Threshold $\alpha$. }We use a hyper-parameter $\alpha$ as the threshold to construct the region-attribute corresponding table in LLaFS (See Sec.3.2.2 and Eq.1 of main paper for details). Fig.\ref{alpha} presents the results when using different values as $\alpha$. It can be observed that both excessively small and large values of $\alpha$ can decrease the mIoU. This could be because an overly small $\alpha$ may result in the incorrect region-attribute match, while an excessively large $\alpha$ may lead to missed matches, both of which can adversely affect the quality of the generated table. When $0.20\leq\alpha\leq0.22$, the model can consistently achieve stable and high performance. 

\vspace{-0.3\baselineskip}
\section{Examples of Key Steps from Input to Output}\label{example_step}
\vspace{-0.3\baselineskip}
We provide two examples to show the key steps from input to output in LLaFS, including network input, class attributes, region-attribute corresponding table, instruction refinement, complete instruction, LLM output, and final result. These examples are presented in the following 5 pages. 

\begin{figure*}[t]
    \centering
    \includegraphics[width=1\linewidth]{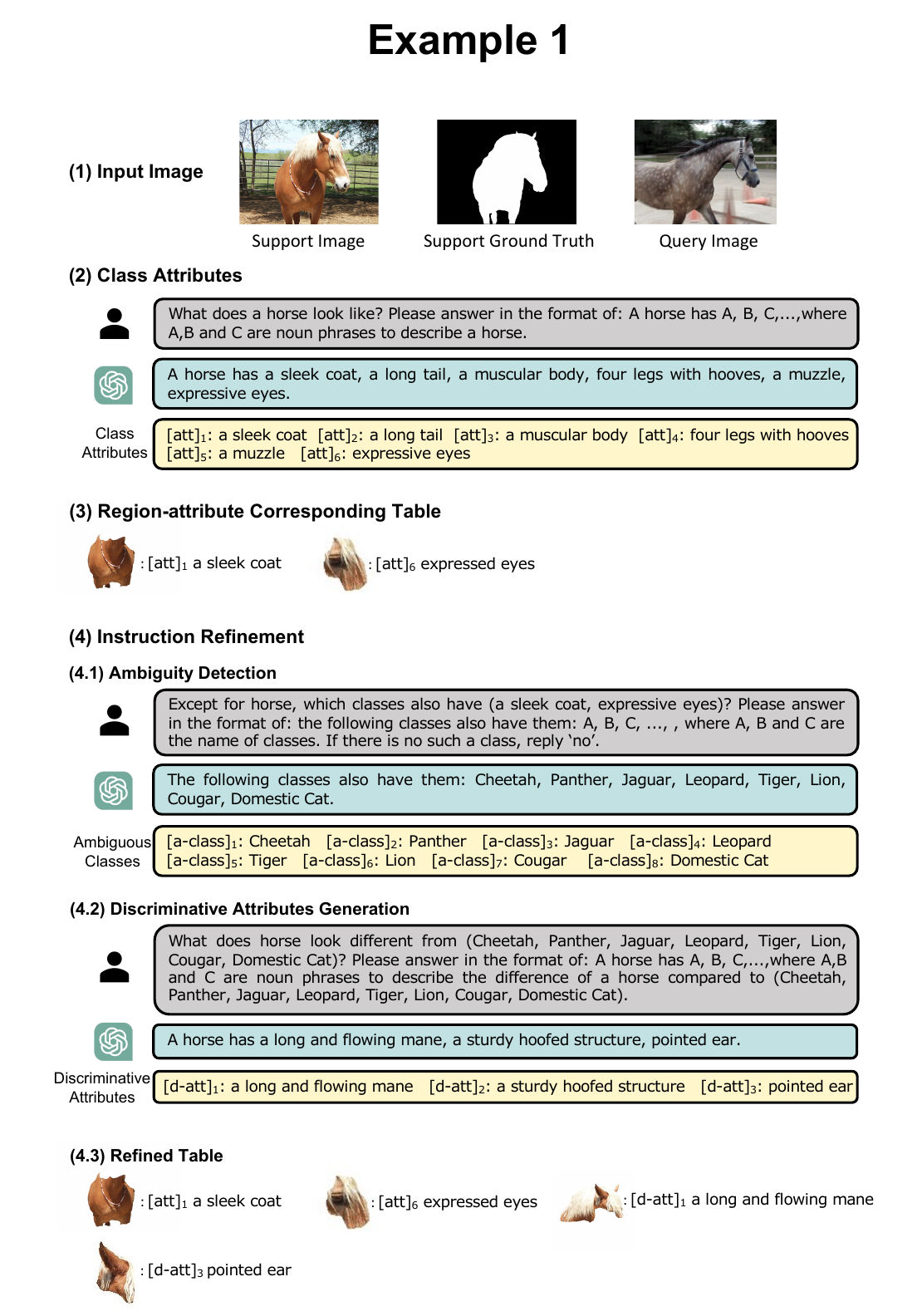}
    \label{all_1}
\end{figure*}

\begin{figure*}[t]
    \centering
    \includegraphics[width=1\linewidth]{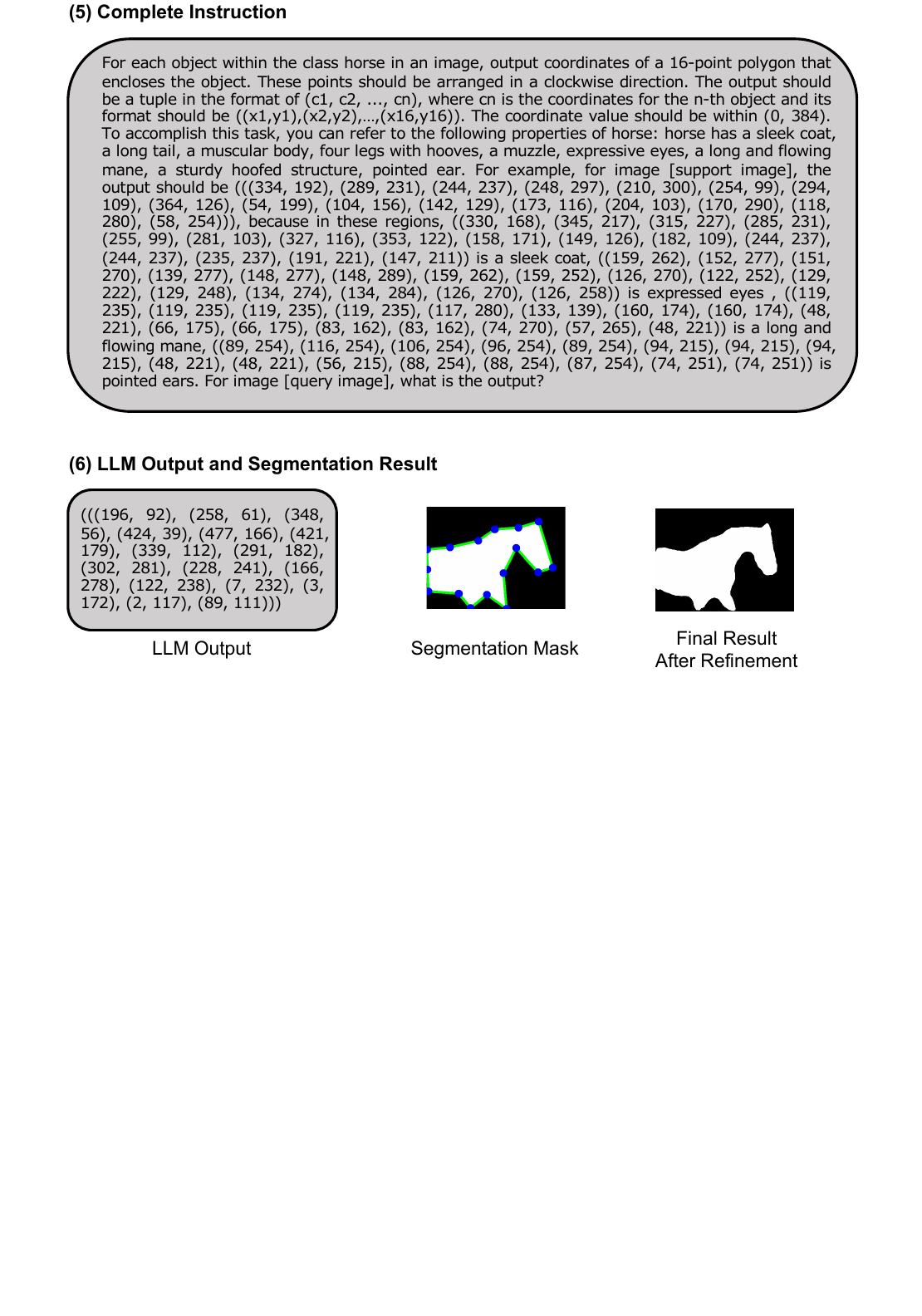}
    \label{all_2}
\end{figure*}

\begin{figure*}[t]
    \centering
    \includegraphics[width=1\linewidth]{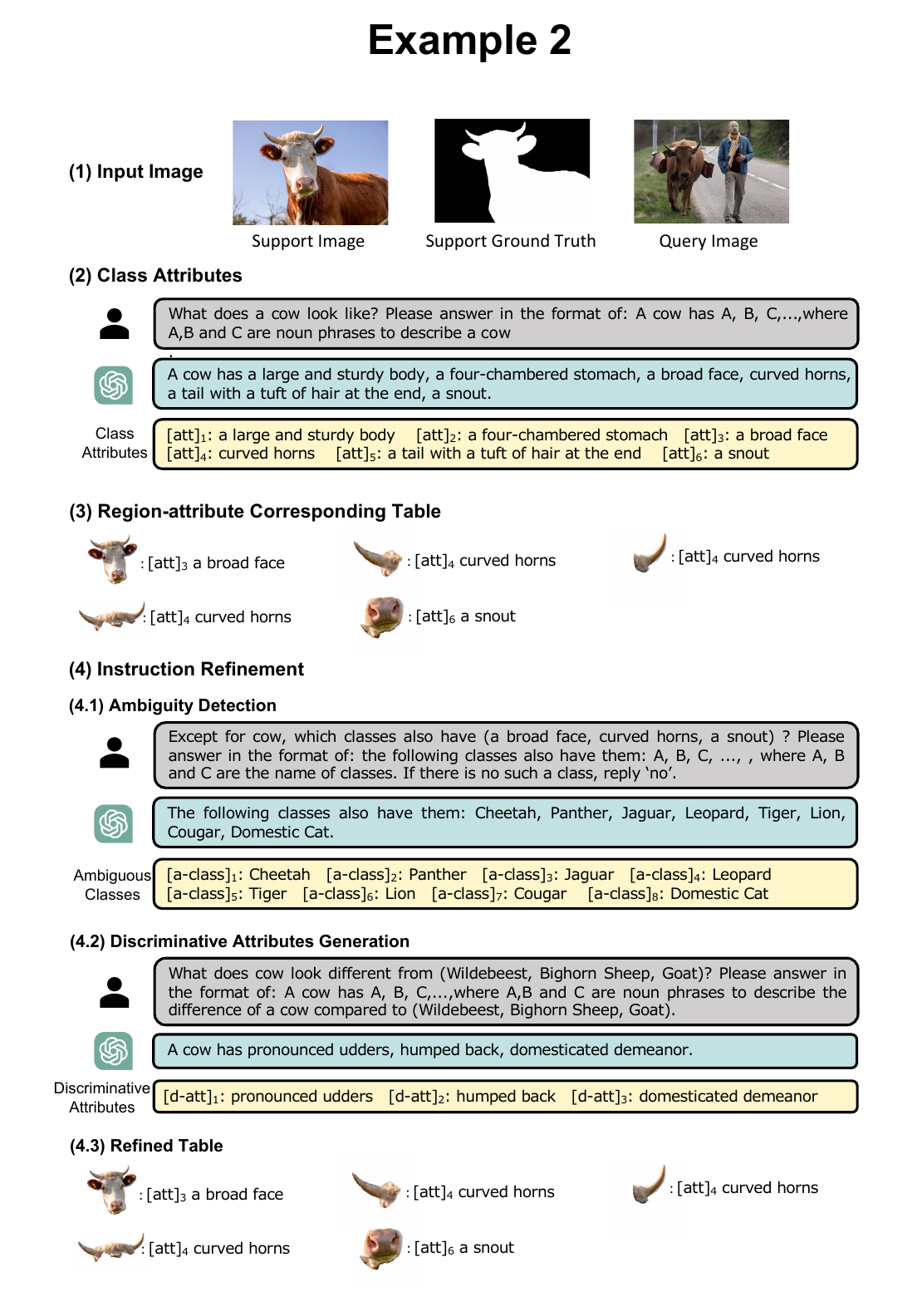}
    \label{all_3}
\end{figure*}

\begin{figure*}[t]
    \centering
    \includegraphics[width=1\linewidth]{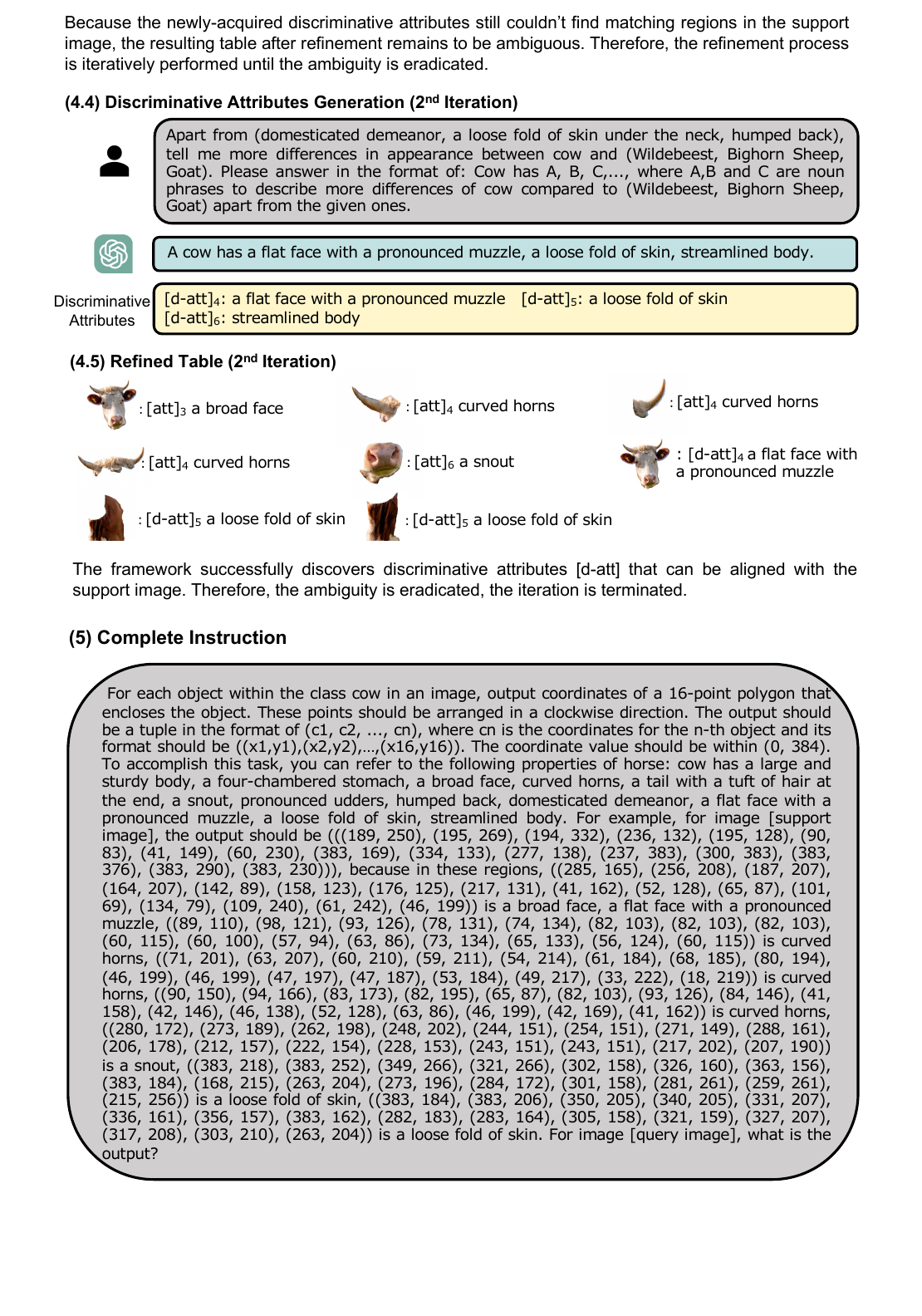}
    \label{all_4}
\end{figure*}

\begin{figure*}[t]
    \centering
    \includegraphics[width=1\linewidth]{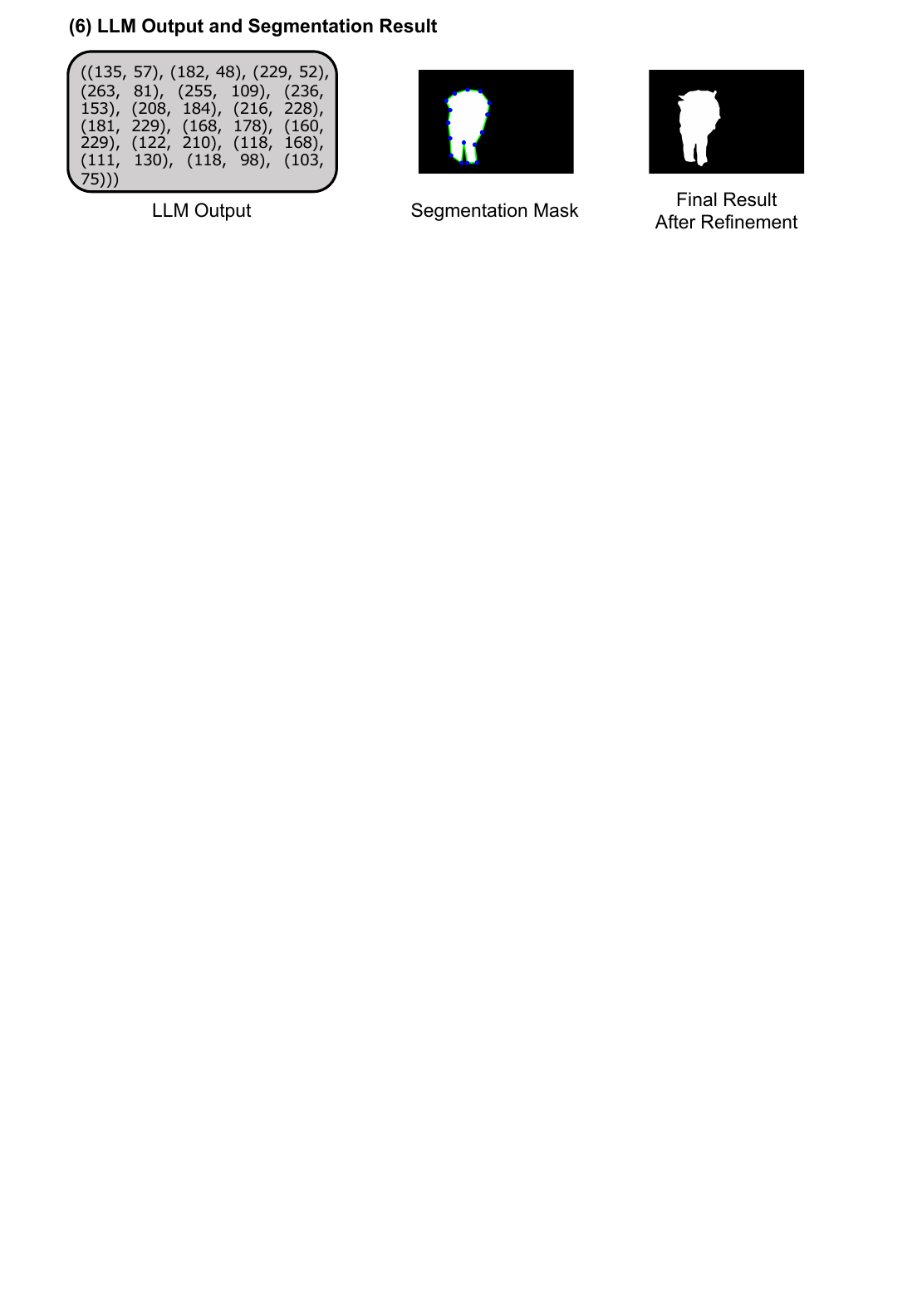}
    \label{all_5}
\end{figure*}


\end{document}